\documentclass{article}
\usepackage[latin9]{inputenc}
\usepackage{array}
\usepackage{float}
\usepackage{textcomp}
\usepackage{multirow}
\usepackage{amsmath,amsfonts,amssymb}
\usepackage{graphicx}
\usepackage{microtype}
\usepackage{xcolor}
\usepackage{amsthm}
\usepackage{bbm}
\usepackage{cuted}
\usepackage{nidanfloat}
\newtheorem{theorem}{Theorem}
\newtheorem{lemma}[theorem]{Lemma}
\newtheorem{definition}{Definition}
\usepackage[unicode=true,
 bookmarks=false,
 breaklinks=false,pdfborder={0 0 1},colorlinks=false]
 {hyperref}

\makeatletter

\providecommand{\tabularnewline}{\\}
\floatstyle{ruled}
\newfloat{algorithm}{tbp}{loa}
\providecommand{\algorithmname}{Algorithm}
\floatname{algorithm}{\protect\algorithmname}

\usepackage{subfigure}
\usepackage[accepted]{icml2021}

\icmltitlerunning{SparseBERT: Rethinking the Importance Analysis in Self-attention}

\def\ff{\text{ff}}

\def\BERT{\text{BERT}}

\begin{document}
\twocolumn[ \icmltitle{SparseBERT: Rethinking the Importance Analysis in Self-attention}
% It is OKAY to include author information, even for blind % submissions: the style file will automatically remove it for you % unless you've provided the [accepted] option to the icml2021 % package.
% List of affiliations: The first argument should be a (short) % identifier you will use later to specify author affiliations % Academic affiliations should list Department, University, City, Region, Country % Industry affiliations should list Company, City, Region, Country
% You can specify symbols, otherwise they are numbered in order. % Ideally, you should not use this facility. Affiliations will be numbered % in order of appearance and this is the preferred way. \icmlsetsymbol{equal}{*}
\begin{icmlauthorlist} \icmlauthor{Han Shi}{hkust} \icmlauthor{Jiahui Gao}{hku} \icmlauthor{Xiaozhe Ren}{noah} \icmlauthor{Hang Xu}{noah} \icmlauthor{Xiaodan Liang}{zs} \icmlauthor{Zhenguo Li}{noah} \icmlauthor{James T. Kwok}{hkust} \end{icmlauthorlist}
\icmlaffiliation{hkust}{Hong Kong University of Science and Technology, Hong Kong} \icmlaffiliation{hku}{The University of Hong Kong, Hong Kong}
\icmlaffiliation{zs}{Sun Yat-sen University, China}
\icmlaffiliation{noah}{Huawei Noah's Ark Lab}
\icmlcorrespondingauthor{Han Shi}{hshiac@cse.ust.hk}
% \icmlcorrespondingauthor{Eee Pppp}{ep@eden.co.uk}
% You may provide any keywords that you % find helpful for describing your paper; these are used to populate % the "keywords" metadata in the PDF but will not be shown in the document \icmlkeywords{Machine Learning, ICML}
\vskip 0.3in ]



\printAffiliationsAndNotice{}  % leave blank if no need to mention equal contribution
\begin{abstract}
Transformer-based models are popularly used in natural language processing
(NLP).
Its core component, self-attention,
has aroused widespread interest.
To understand the self-attention mechanism,
a direct method
is to visualize
the attention map
of a pre-trained model.
Based on the
patterns observed,
a series of efficient Transformers
with different sparse attention masks
have been proposed.
From a theoretical perspective,
universal approximability of Transformer-based models is also recently proved.
However, the above understanding and analysis of self-attention is based on a pre-trained model. To rethink the importance analysis
in self-attention, we study the significance of different positions in attention matrix during
pre-training. A surprising result
is that diagonal elements in the attention map are the least important
compared with other attention positions. We provide a proof showing that
these diagonal elements can indeed be removed without deteriorating model performance.
Furthermore, we propose a Differentiable Attention Mask (DAM) algorithm, which
further guides the design of the SparseBERT.
Extensive experiments verify our interesting findings and illustrate the effect
of the proposed algorithm.
\end{abstract}

\section{Introduction}

The Transformer \cite{vaswani2017attention} has been commonly used
in various natural language processing (NLP) tasks such as
text classification \cite{wang2018glue}, text translation \cite{ott2018scaling},
and
question answering \cite{rajpurkar2016squad}. The
recent use of Transformer for image classification \cite{dosovitskiy2020image}, object detection \cite{carion2020end}
also demonstrates its potential in computer vision.
Two notable descendants from
the Transformer include the
BERT \cite{devlin2019bert},
which
achieves state-of-the-art performance
on a wide range of NLP tasks, and
GPT-3 \cite{brown2020language} which applies the
Transformer's
decoder
on generative downstream tasks.

Self-attention is a core component in Transformer-based architectures.
Recently, its interpretation has aroused a lot of interest.
Visualization has been commonly used to understand the attention map during inference \cite{park2019sanvis,gong2019efficient,kovaleva2019revealing}. For example,
\citeauthor{park2019sanvis}
(\citeyear{park2019sanvis})
and
\citeauthor{gong2019efficient}
(\citeyear{gong2019efficient})
randomly select a sentence from the corpus and visualize the attention maps of
different heads in a pre-trained Transformer model.
\citeauthor{kovaleva2019revealing}
(\citeyear{kovaleva2019revealing})
summarizes five attention
patterns and estimates their ratios in different tasks.
A common observation
from these studies
is that
local attention and global attention are both important for
token understanding.

While self-attention is powerful, a  main concern
is its efficiency bottleneck. As each token
has to attend to all $n$ tokens
in the sequence,
the complexity scales as $\mathcal{O}(n^{2})$.
This can be
expensive
on long sequences.
To alleviate this problem, sparse attention allows each token to attend to only a token subset.
A series of Transformer variants have been proposed along this direction
\cite{guo2019star,child2019generating,li2019enhancing,beltagy2020longformer}.
However, these sparse attention schemes are designed manually.
It is still an open issue
on how to find a suitable attention scheme.

\begin{figure*}[t]
\centering
\subfigure[Star.\label{subfig:star}]{\includegraphics[width=0.16\textwidth]{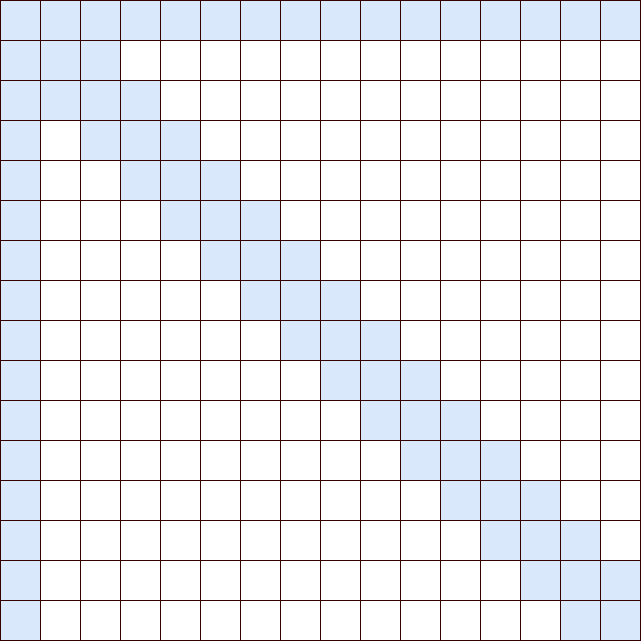}}
\subfigure[LogSparse.\label{subfig:logsparse}]{\includegraphics[width=0.16\textwidth]{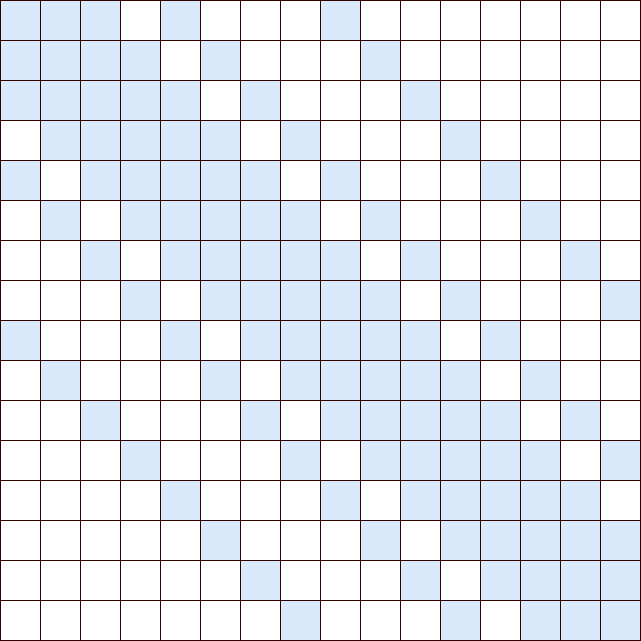}}
\subfigure[Strided.\label{subfig:strided}]{\includegraphics[width=0.16\textwidth]{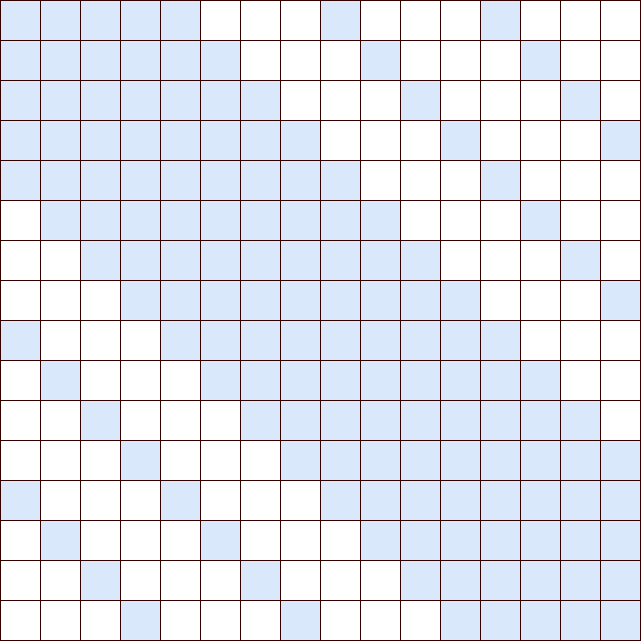}}
\subfigure[Fixed.\label{subfig:fixed}]{\includegraphics[width=0.16\textwidth]{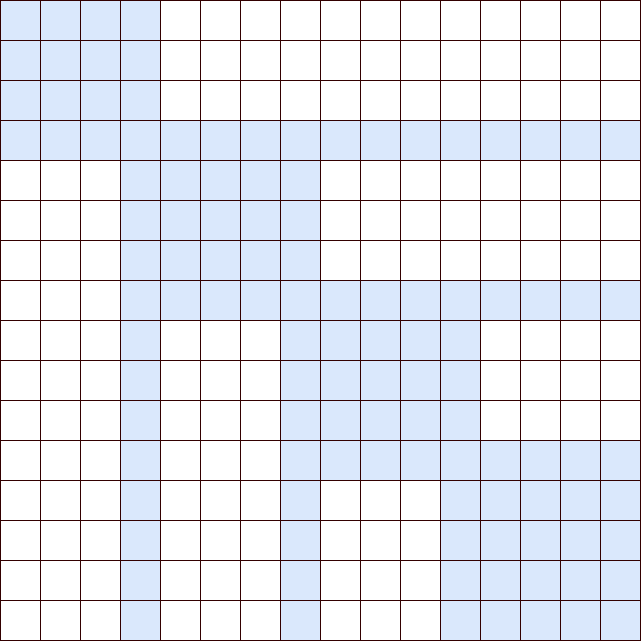}}
\subfigure[Longformer.\label{subfig:long}]{\includegraphics[width=0.16\textwidth]{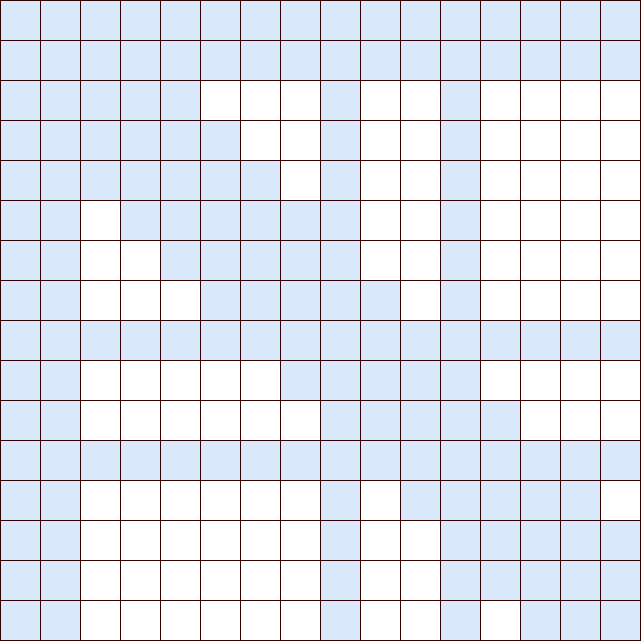}}
\subfigure[BigBird.\label{subfig:bigbird}]{\includegraphics[width=0.16\textwidth]{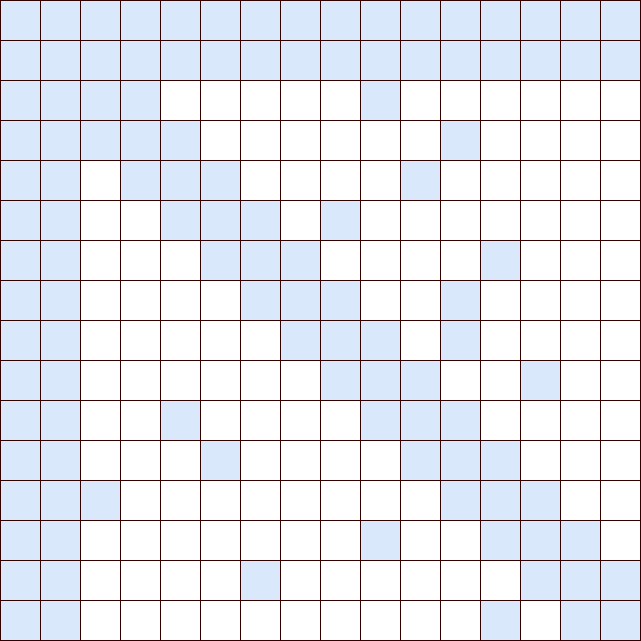}}
\caption{Examples of existing attention masks (with $n=16$). \label{fig:The-visualization-of-1}}
\end{figure*}

Recently,  there is a growing interest in
understanding self-attention mechanism from a theoretical perspective. Results
show that the Transformer and its variants are universal approximators of arbitrary continuous sequence-to-sequence functions \cite{yun2019Transformers,yun2020n,zaheer2020big}.
A key part of their proofs is that self-attention layers
implement contextual mappings of the input sequences.
\citeauthor{yun2019Transformers}
(\citeyear{yun2019Transformers})
constructs the self-attention model as a selective shift operation
such that contextual mapping can be implemented.
\citeauthor{zaheer2020big}
(\citeyear{zaheer2020big})
shows that universal approximation holds for
their sparse Transformer BigBird
if its attention structure contains the star graph.
\citeauthor{yun2020n}
(\citeyear{yun2020n})
provides a unifying framework for the universal approximation of sparse Transformers.
Note that they all emphasize the
importance of diagonal elements in the attention map.

To guide the design of an efficient Transformer,
it is useful to investigate the importance of different positions
in self-attention.
In this paper, we study
this using differentiable
search \citep{liu2018darts,xie2018snas}.
A learnable attention distribution
is constructed
and
the score of each position
is learned
in an end-to-end manner during pre-training.
While existing theoretical and empirical findings suggest the importance of diagonal elements
in the self-attention matrix, we observe that they are indeed the least important
compared to other entries. Furthermore, neighborhood tokens and special
tokens (such as the first token {\tt [CLS]} and last token {\tt [SEP]}) are also
prominent, which is consistent with previous observations in
\citep{park2019sanvis,gong2019efficient,kovaleva2019revealing,clark2019does}.
Besides,
using the Gumbel-sigmoid function \citep{maddison2017concrete},
we propose the Differentiable Attention Mask (DAM)
algorithm to learn the attention mask in an end-to-end manner.
Extensive experiments using
masks with different sparsity ratios
on various
NLP tasks
demonstrate the effect of the proposed algorithm.
Specifically, highly sparse structured attention masks (with $91.3\%$ sparsity ratio)
can already achieve
$80.9\%$ average score on the GLUE development set \cite{wang2018glue}. The code is available at \url{https://github.com/han-shi/SparseBERT}.

\section{Related Work}

\subsection{Transformer Block and Self-Attention \label{sec:trans_block}}

The Transformer block is a basic component in the Transformer \cite{vaswani2017attention}
and BERT \cite{devlin2019bert} architectures.
Let $\boldsymbol{X}\in\mathbb{R}^{n\times d}$ be the input to a Transformer block, where $n$ is the number of input tokens and $d$ is the embedding size.
Each block consists of a self-attention layer and a feed-forward layer.

The self-attention
layer output
can be written
as:
\begin{align}
Attn(\boldsymbol{X}) &=  \boldsymbol{X}+\sum_{k=1}^{H}\sigma(\boldsymbol{X}\boldsymbol{W}_{Q}^{k}(\boldsymbol{X}\boldsymbol{W}_{K}^{k})^{\top} )\boldsymbol{XW}_{V}^{k}\boldsymbol{W}_{O}^{k\top} \nonumber\\
&=\boldsymbol{X}+\sum_{k=1}^{H}\boldsymbol{A}^{k}(\boldsymbol{X})\boldsymbol{V}^{k}(\boldsymbol{X})\boldsymbol{W}_{O}^{k\top},  \label{eq:attn}
\end{align}
where $H$ is the number of heads,
$\sigma$ is
the softmax function, and
$\boldsymbol{W}_{Q}^{k},\boldsymbol{W}_{K}^{k},\boldsymbol{W}_{V}^{k},\boldsymbol{W}_{O}^{k}\in\mathbb{R}^{d\times
d_{h}}$ (where $d_{h}=d/H$ is the dimension of a single-head output) are weight matrices for the query, key, value, and output,
respectively of the $k$th head. In particular,
the self-attention matrix
\begin{equation} \label{eq:matrix}
\boldsymbol{A}(\boldsymbol{X})=\sigma(\boldsymbol{XW}_{Q}(\boldsymbol{XW}_{K})^{\top})
\end{equation}
    in (\ref{eq:attn})
plays a key role in the self-attention layer
\cite{park2019sanvis,gong2019efficient,kovaleva2019revealing}.

The fully-connected layer usually has
two layers
with residual connection:
{\small
\[
FF(\boldsymbol{X})=Attn(\boldsymbol{X})+ReLU(Attn(\boldsymbol{X})\boldsymbol{W}_{1}+\boldsymbol{b}_{1})\boldsymbol{W}_{2}+\boldsymbol{b}_{2},
\]}where
$\boldsymbol{W}_{1}\in\mathbb{R}^{d\times
d_{\ff}},\boldsymbol{W}_{2}\in\mathbb{R}^{d_{\ff}\times d}$ ($d_{\ff}$ is the size of the intermediate layer)
are the weight matrices, and
$\boldsymbol{b}_{1},\boldsymbol{b}_{2}$
are the biases.
As in \cite{yun2019Transformers,zaheer2020big,yun2020n}, we drop the scale
product and layer-normalization layer
to simplify analysis.

\subsection{Sparse Transformers\label{subsec:Efficient-Transformer}}

To reduce the quadratic complexity in self-attention,
a number of
sparse Transformers
have been
recently
proposed.
In these models, each token
can attend to only a subset of fixed positions
\cite{guo2019star,child2019generating,li2019enhancing,beltagy2020longformer}.
This can be seen as being controlled by an attention mask
$\boldsymbol{M}=[0,1]^{n\times n}$, where
$M_{i,j}=1$ indicates that token $i$ can attend to token $j$, and $0$ otherwise.
For example,
Figure~\ref{subfig:star}
shows
the attention mask of
the Star-Transformer
\cite{guo2019star}.
It uses ring connections for
local attention,
and radical connections to an auxiliary relay node
(the first token in the figure)
to represent global attention.
\citeauthor{li2019enhancing}
(\citeyear{li2019enhancing})
proposes the LogSparse self-attention, in which each token only attends to
itself and its previous tokens with an exponential stepsize
(Figure~\ref{subfig:logsparse}),
resulting in $\mathcal{O}(n\log n)$ complexity.
\citeauthor{child2019generating}
(\citeyear{child2019generating})
performs sparse factorization
on the attention matrix, and
reduces its complexity to $\mathcal{O}(n\sqrt{n})$ with the use of
two attention masks.
The strided mask
(Figure~\ref{subfig:strided})
attends to every $l$th location
(where $l$ is the stride step),
while the fixed mask
(Figure~\ref{subfig:fixed})
allows specific positions to be attended to.
The very recent Longformer \cite{beltagy2020longformer} (Figure~\ref{subfig:long}) and
BigBird \cite{zaheer2020big} (Figure~\ref{subfig:bigbird}) models use a number of attention patterns,
and reduce their complexities to
$\mathcal{O}(n)$. With these sparse Transformers,
BERT is shown to be more efficient for long document understanding
\cite{child2019generating,qiu2020blockwise}.

\section{Which Attention Positions are Important?\label{sec:Which-attention-position}}

Previous works on sparse Transformers only provide a crude understanding of the self-attention module that local attention and global attention
are both important.
In this section, we study the self-attention matrix
$\boldsymbol{A}\in\mathbb{R}^{n\times n}$ in
Eq. (\ref{eq:matrix})
in more detail.
To emphasize its role,
we write the output of the self-attention layer as $Attn(\boldsymbol{X}, \boldsymbol{A}(\boldsymbol{X}, \boldsymbol{M}))$, where $\boldsymbol{M}$ is a fixed attention mask.
Since the nonzero elements of the attention matrix are fixed, one
only needs to perform computations related to these positions.
We define
the sparsity of an attention mask
as $\rho=1-|\boldsymbol{M}|/n^{2}$.
The complexity of the self-attention layer
is thus reduced to
$\mathcal{O}((1-\rho) n^2)$.

With a low self-attention sparsity, a token can attend to more tokens given the same amount of computational cost, and
is thus expected to have better
performance.
On the other hand,
at high self-attention sparsity,
model performance may drop.
It is natural to ask which positions in self-attention are
more important.
In other words, which attention mask is better for a given sparsity.
We formulate this problem as the search  for
a mask
in $[0,1]^{n\times n}$
such that the balance between performance and efficiency is optimized.

\subsection{Continuous Relaxation}
\label{subsec:Continuous-Relaxation}

In this section, we investigate
the importance of different positions in self-attention.
A similar study on the importance of different heads in the Transformer is recently performed in
\cite{michel2019sixteen}.
However,
while \citeauthor{michel2019sixteen}
(\citeyear{michel2019sixteen})
performs the ablation study
with only $16$ heads, there are
$2^{n\times n}$
possible attention distributions
here. This huge search space makes the study very challenging.

In neural architecture search (NAS) \citep{elsken2019neural},
one has to find a good architecture from a huge search space.
Inspired by the similarity with our problem,
we propose to
use continuous relaxation
as in differentiable architecture search (DARTS) \citep{liu2018darts}.
Specifically, we associate an
$\alpha_{i,j}$ with each
position $(i,j)$ in the self-attention matrix $\boldsymbol{A}(\boldsymbol{X})$,
and define
the attention probability
as
\begin{equation} \label{eq:P}
P_{i,j}=\text{sigmoid}(\alpha_{i,j}) \in [0,1].
\end{equation}
For symmetric structure, we enforce
$\alpha_{i,j}=\alpha_{j,i}$.
Analogous to Eq. (\ref{eq:attn}),
the soft-masked self-attention is then
{\small
\begin{align}
Attn(\boldsymbol{X})=\boldsymbol{X}+\sum_{k=1}^{H} (\boldsymbol{P}^k\odot \boldsymbol{A}^{k}(\boldsymbol{X}))\boldsymbol{V}^{k}(\boldsymbol{X})\boldsymbol{W}_{O}^{k\top}, \label{eq:modified}
\end{align}}where $\odot$ is the element-wise product. Obviously, when $P_{i,j}=1$ for all $(i,j)$'s,
this reduces to Eq. (\ref{eq:attn}). However, the above multiplicative attention
mask will result in unnormalized attention distributions. To solve this problem,
we introduce the renormalization trick, which replaces the multiplicative attention mask with an additive mask before the softmax function as follows.
\begin{align}
\boldsymbol{\hat{A}}(\boldsymbol{X})&=\sigma(\boldsymbol{XW}_{Q}(\boldsymbol{XW}_{K})^{\top}+\boldsymbol{Q}),\\
Q_{i,j}&=-c(1-P_{i,j}),
\end{align}
where $\boldsymbol{Q}\in \mathbb{R}^{n\times n}$ is the addictive attention mask,
$c$ is a large constant such that $\hat{A}_{i,j}=0$ if $P_{i,j}=0$, and
$\hat{A}_{i,j}$ reduces to the original attention score if $P_{i,j}=1$.

As in DARTS,
there are two sets of learnable parameters:
parameter $\boldsymbol{w}$ of the Transformer model, and the attention parameter
$\boldsymbol{\alpha}=\{ \alpha_{i,j}\}$.
They can be learned by using
either one-level optimization \cite{xie2018snas} or bi-level optimization
\cite{liu2018darts}
formulations.
Recently,
\citeauthor{bi2020gold}
(\citeyear{bi2020gold})
shows that
the limitations of one-level optimization
can be alleviated
when a large data set
is used.
In our context,
as a large data set is often available
during pre-training,
we apply the simpler one-level optimization here.

\subsubsection{Experimental Setup}
In this experiment,
we empirically study the effect of different positions in the self-attention module
using the BERT-base.
This model is
stacked with $12$ Transformer blocks (Section~\ref{sec:trans_block}) with the
following hyper-parameters: number of tokens $n=128$, number of self-attention heads $h=12$, and hidden layer size $d=768$.
For better comparison with prior works in Figure~\ref{fig:The-visualization-of-1},
the parameters of $\boldsymbol{P}$
are also shared among blocks, leading to a total of $12$ $\boldsymbol{P}$'s (one for each
self-attention head).
As for the feed-forward layer, we set
the filter size $d_{\ff}$ to 3072 as in \cite{devlin2019bert}.
We follow the standard pre-training
experiment setting in \cite{devlin2019bert}, and take Masked Language Modeling (MLM) and Next Sentence Prediction (NSP) as pre-training tasks.
Data sets
BooksCorpus
(with $800$M words) \cite{zhu2015aligning} and English Wikipedia (with $2,500$M
words) \cite{devlin2019bert} are  used.
We use the WordPiece embedding \citep{wu2016google}, and $30,000$ tokens are
contained in the dictionary. The special token {\tt [CLS]} is used as the first
token of each sequence. Another special token {\tt [SEP]} is used to separate
sentences in a sequence.
The pre-training is performed for $40$ epochs.
All experiments
are performed on NVIDIA Tesla V100 GPUs.

\subsubsection{Results}
\label{sec:visual}

\begin{figure}[t]
\centering
\subfigure[Attention distribution.
\label{subfig:exp_res}]{\includegraphics[width=0.24\textwidth]{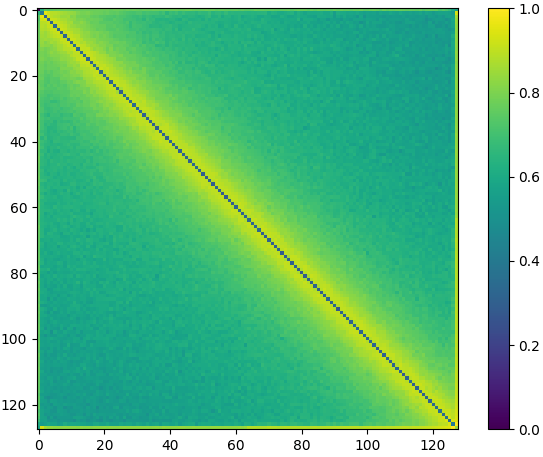}}
\subfigure[Sketch.\label{subfig:sketch}]{\includegraphics[width=0.21\textwidth]{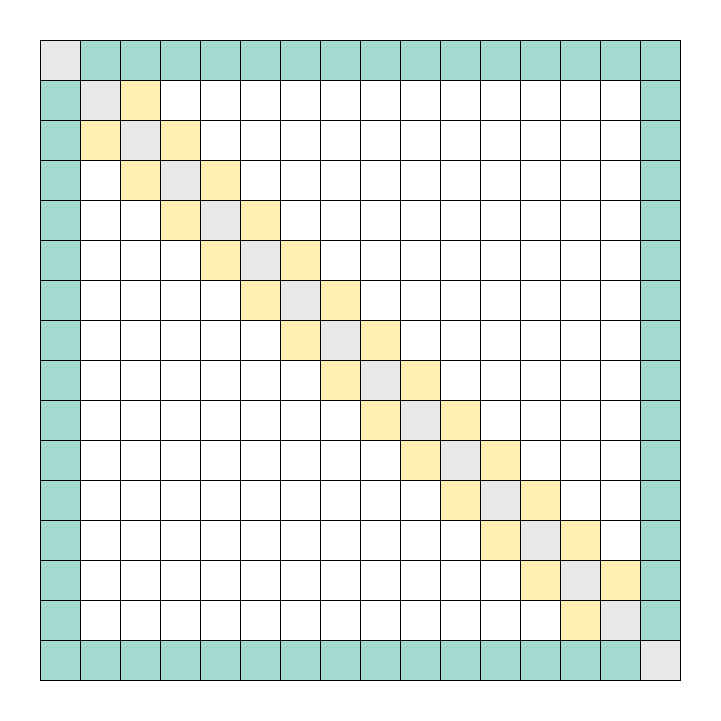}}
\caption{Visualization of the attention distribution. In
Figure~\ref{subfig:sketch},
the dark entries are for diag-attention, yellow for neighborhood attention
and green for attention to special tokens. \label{fig:importance_sc}}
\end{figure}

Figure~\ref{subfig:exp_res}
shows the
attention distribution $\boldsymbol{P}$ averaged over the $12$ heads after normalization (with $n=128$). For comparison with existing attention masks in Figure~\ref{fig:The-visualization-of-1}, we also illustrate its sketch in Figure~\ref{subfig:sketch} (with $n=16$).
The following can be observed: (i)
Diagonal elements
(denoted ``diag-attention" in the sequel)
are the least important compared to other positions.
Surprisingly, this conflicts with existing
observations in \cite{park2019sanvis,gong2019efficient,kovaleva2019revealing}, which emphasize the importance of diagonal attention.
We believe this is because the self-attention layer
already has
a skip connection (the term $\boldsymbol{X}$
in (\ref{eq:attn})).
Influence of diag-attention can thus be conveyed in the skip connection
instead of via the self-attention matrix;
(ii) Neighborhood positions are the most significant in the attention distribution
matrix; (iii) The special tokens ({\tt [CLS]} and {\tt [SEP]}) are important,
which is also observed in \cite{clark2019does}; (iv)
The importance of other positions
is similar.

\subsection{Universal Approximability}

Recall that the Transformer and its variants are universal approximators of arbitrary continuous sequence-to-sequence functions \cite{yun2019Transformers,yun2020n,zaheer2020big}.
In their proofs,
diagonal positions in the self-attention matrix play a key role.
As Section~\ref{subsec:Continuous-Relaxation} has shown that the diagonal
elements are empirically the least important,
an interesting question is
whether
universal approximability will still hold when
these diagonal elements
(diag-attention)
are dropped.

Without diag-attention, the
$i$th token
output of
the self-attention
layer becomes:
\[
Attn(\boldsymbol{X})_{i}=\boldsymbol{X}_{i}+\sum_{k=1}^{H}\sum_{j\neq i}A_{i,j}^{k}(\boldsymbol{X})\boldsymbol{V}_{j}^{k}(\boldsymbol{X})\boldsymbol{W}_{O}^{k\top}.
\]
Let
$\mathcal{T}^{H,d_h,d_{\ff}}$ be a class of Transformers without diag-attention
stacks, and
$F_{CD}$ be the set of continuous functions $f:[0, 1]^{n\times d}\mapsto
{\mathbb R}^{n\times d}$. For any $p\geq 1$, the $\ell_p$-distance between
$f_1, f_2 \in
F_{CD}$
is defined as $d_p(f_1, f_2)=(\int\|f_1(\boldsymbol{X})-f_2(\boldsymbol{X})\|^p_pd\boldsymbol{X})^{1/p}$.
The following Theorem shows that the self-attention mechanism without diag-attention is also a universal approximator:
\begin{theorem}
Given $1<p<\text{\ensuremath{\infty}}$, $\epsilon>0$ and $n>2$, for any
$f\in F_{CD}$, there exists a Transformer network without diag-attention $g\in
T^{2,1,4}$,
such that $d_{p}(f,g)<\epsilon$.
\end{theorem}

The following shows
the proof outline, which is similar to that in \cite{yun2019Transformers}. The
main difference is in the contextual mapping step since each token cannot
attend to itself in our scenario.

\textbf{Step 1}: Approximate $F_{CD}$ with the set of piecewise-constant functions
$\bar{F}_{CD}$. We split input $[0, 1]^{n\times d}$ into a set of grids
$\mathcal{G}_\delta\in\{0, \delta, \dots, 1\}^{n\times d}$. We then approximate
any input belonging to the same cube $\mathcal{G}_\delta+[0, \delta]^{n\times d}$
by the same value, resulting in a piecewise-constant function $\bar{f}$. With
$\delta$
small enough,
we have $d_p(f, \bar{f}) \leq \epsilon/3$.

\textbf{Step 2}: Approximate $\bar{F}_{CD}$ with the modified Transformer
blocks $\bar{\mathcal{T}}^{H,d_h,d_{\ff}}$, which replace the softmax operator and ReLU with the hardmax operator and a piece-wise linear functions (at most three pieces). For each above $\bar{f}$, there exists a closely approximate function $\bar{g}\in \bar{\mathcal{T}}^{2,1,1}$ such that $d_p(\bar{f},\bar{g})=O(\delta^{d/p})$.

This is the key step related to the contextual mapping. A selective
shift operation is proposed to construct an approximation in \cite{yun2019Transformers}.
Here, we consider a simple scenario where $n=3$ and $d=1$
and let
$\boldsymbol{L}=[l_{1}\quad l_{2}\quad l_{3}]^{\top}\in\mathcal{G}_{\delta}$.
Without loss of generality, we assume that $l_{1}<l_{2}<l_{3}$.
For a Transformer without diag-attention, the selective shift operation,
consisting of $2$ attention heads of size $1$, is constrained as follows:
\begin{align*}
\lefteqn{\Psi(\boldsymbol{Z};b_{1},b_{2})_{i,1}} \\
&& = \left\{
\begin{array}{ll}
\max_{j\neq i}Z_{j,1}\text{\textminus}\min_{j\neq i}Z_{j,1}
& \text{if}\;b_{1}<Z_{i,1}<b_{2} \\
0
& \text{otherwise}
\end{array} \right..
\end{align*}
We stack $1/\ensuremath{\delta}$ self-attention layers, with attention
parts $\delta^{-1}\Psi(\text{\ensuremath{\cdot}};l\text{\textminus}\delta/2,l+\delta/2)$
for each $l\in\{0,\delta,\dots,1-\delta\}$ in increasing order of
$l$.  The
shift operation is first applied to $l_{1}$, resulting in $\widetilde{l_{1}}=l_{1}+\delta^{-1}(l_{3}-l_{2})>l_{3}$.
The second shift operation is
then
applied to the second element, resulting in $\widetilde{l_{2}}=l_{2}+\delta^{-1}(\widetilde{l_{1}}-l_{3})=l_{2}+\delta^{-1}(l_{1}-l_{3})+\delta^{-2}(l_{3}-l_{2})>\widetilde{l_{1}}.$
Finally, a similar operation is applied to $l_{3}$, and
the shifted result is $\widetilde{l_{3}}=l_{3}+\delta^{-1}(\widetilde{l_{2}}-\widetilde{l_{1}})=l_{3}+\delta^{-1}(l_{2}-l_{1})+\delta^{-2}(l_{1}+l_{2}-2l_{3})+\delta^{-3}(l_{3}-l_{2})$.
It is easy to check that the map from the original $\boldsymbol{L}$ to $\widetilde{l_{3}}$
is one-to-one and that $0<\widetilde{l_{1}}<\widetilde{l_{2}}<\widetilde{l_{3}}<\delta^{-3}$.
We then add two additional layers shifting all positive elements, resulting in
$[\widetilde{l_{1}}+\Delta(\widetilde{l_{2}}+\widetilde{l_{3}})+\Delta^2\widetilde{l}_3 \quad \widetilde{l_{2}}+\Delta(\widetilde{l_{1}}+\widetilde{l_{3}})+\Delta^2\widetilde{l}_3 \quad \widetilde{l_{3}}+2\Delta\widetilde{l_{2}}+\Delta^2\widetilde{l}_3]^{\top}$, where $\Delta=(\delta^{-1}-1)(\delta^{-3}+\delta^{-1}+1)$.
Note that all elements are in the disjoint interval
for different $\boldsymbol{L}$'s because $\boldsymbol{L}\rightarrow\widetilde{l_{3}}$ is bijective.
Thus, the
self-attention layer without diag-attention is a
contextual mapping as defined in \cite{yun2019Transformers}.

\begin{table*}[t]
\caption{Performance
(in \%)
of the various
BERT-base variants
on the GLUE data set. \label{tab:glue}}
\centering
\resizebox{0.99\textwidth}{!}{
\begin{tabular}{lccccccccc}
\hline
 & \multirow{1}{*}{MNLI (m/mm)} & QQP & QNLI & SST-2 & COLA & STS-B & MRPC & RTE & Average\tabularnewline
\hline
\multicolumn{10}{l}{\textbf{\textit{Development Set}}} \tabularnewline
BERT-base (ours) & 85.4/85.8 & 88.2 & 91.5 & 92.9 & 62.1 & 88.8 & 90.4 & 69.0 & 83.8 \tabularnewline
BERT-base (randomly dropped) & 84.6/85.2 & 87.7 & 91.1 & 92.7 & 62.0 &  88.9 & 89.3& 68.9  & 83.4 \tabularnewline
\multirow{1}{*}{BERT-base (no diag-attention)} & 85.6/85.9 & 88.2 & 92.0 & 93.8 & 63.1 & 89.2 & 91.2 & 67.9 & \textbf{83.9} \tabularnewline
\hline
\multicolumn{10}{l}{\textbf{\textit{Test Set}}} \tabularnewline
BERT-base \citep{devlin2019bert} & 84.6/83.4 & 71.2 & 90.5 & 93.5 & 52.1 & 85.8 & 88.9 & 66.4 & 79.6 \tabularnewline
BERT-base (ours) & 84.8/84.1 & 71.3 & 90.9 & 93.4 & 52.3 & 85.3 & 88.3 & 66.9 & 79.7 \tabularnewline
BERT-base (randomly dropped) & 84.5/83.5 & 70.3 & 91.1 & 93.4 & 52.0 & 85.8 & 87.4 & 66.7 & 79.4 \tabularnewline
\multirow{1}{*}{BERT-base (no diag-attention)} & 85.5/84.9 & 71.3 & 91.1 & 93.4 & 53.3 & 86.3 &  88.9 & 67.9 & \textbf{80.3} \tabularnewline
\hline
\end{tabular}}
\end{table*}

\begin{table*}[t]
\begin{center}
\caption{Performance (in \%) of the various BERT-base variants on the SWAG and SQuAD development sets.}
\label{tab:swag}
\centering{}
\begin{tabular}{lccccc}
\hline
& \multicolumn{1}{c}{SWAG} & \multicolumn{2}{c}{SQuAD v1.1} & \multicolumn{2}{c}{SQuAD v2.0} \tabularnewline
  & acc & EM & F1 & EM & F1 \\
\hline
BERT-base \cite{devlin2019bert}   & 81.6 & 80.8 & 88.5 & - & - \tabularnewline
BERT-base (ours)  & 82.5 & 79.7  & 87.1 &  72.9 & 75.5 \tabularnewline
BERT-base (randomly dropped)  & 81.6 & 79.7 & 87.0 & 71.5 & 74.2 \tabularnewline
BERT-base (no diag-attention)  & 83.5 & 80.3 & 87.9 & 73.2 & 75.9 \tabularnewline
\hline
\end{tabular}
\end{center}
\end{table*}

\textbf{Step 3}: Approximate the modified Transformer blocks $\bar{g}\in \bar{\mathcal{T}}^{2,1,1}$ with standard
Transformer blocks $g\in \mathcal{T}^{2,1,4}$ such that we have $d_p(\bar{g},g) \leq \epsilon/3$.

Summarizing the above three steps, we have:
{\small
\[
d_p(f,g)\leq d_p(f,\bar{f})+d_p(\bar{f},\bar{g})+d_p(\bar{g},g)\leq 2\epsilon/3+O(\delta^{d/p}).
\]}With
a small enough
$\delta$, we have $d_p(f,g) \leq \epsilon$.
Thus, Transformers
without diag-attention are also universal approximators. The detailed
proof is in Appendix~\ref{app:proof}.

\subsection{Empirical Verification \label{sec:glue}}

In this section, we empirically study the effect of dropping diag-attention from the self-attention mechanism. The fine-tuning experiments are performed
on the GLUE benchmark \cite{wang2018glue},
SWAG \cite{zellers2018swag} and SQuAD \cite{rajpurkar2016squad,rajpurkar2018know} data sets.

\subsubsection{Data}

The GLUE benchmark
includes
three categories of
natural language understanding tasks:
(i) single-sentence tasks (CoLA and SST-2); (ii) similarity and paraphrase tasks (MRPC, QQP and STS-B); (iii) inference tasks (MNLI, QNLI and RTE).
For MNLI sub-task, we experiment
on both the matched (MNLI-m) and mismatched (MNLI-mm) sections.
The SWAG data set is
for grounded commonsense inference, while the
SQuAD data set
is for question answering.
In SQuAD v1.1, the answers are included in the context; while in SQuAD v2.0, some
answers are not included.
Descriptions of the data sets are in Appendix~\ref{app:dataset}.

\subsubsection{Metric}
Following BERT \cite{devlin2019bert}, we
report different metrics for different
GLUE
sub-tasks. Specifically, we use accuracy
for MNLI, QNLI, RTE, SST-2 tasks, F1 score for QQP and MRPC, Spearman
correlation for STS-B, and Matthews correlation for CoLA.
The results of GLUE test set are evaluated by the evaluation server (\url{https://gluebenchmark.com}).
For SWAG task, we use accuracy for evaluation.
For SQuAD v1.1 and v2.0, we use the Exact Match (EM) and F1 scores.

\subsubsection{Details} We perform pre-training and fine-tuning. The experiment settings
are the same as \citet{devlin2019bert}.
We take the BERT-base model in Section~\ref{subsec:Continuous-Relaxation} as
the baseline and verify the effect of diagonal elements in the self-attention matrix.
In the variant ``BERT-base (no diag-attention)",
the diag-attention in each self-attention layer
is removed, dropping a
total of $128$ elements (as $n=128$).
In ``BERT-base (randomly dropped)", we remove the same number
($128$)
of entries
randomly. To reduce statistical
variability, the results are averaged over three random repetitions.
Details of other hyper-parameter settings are in Appendix~\ref{app:hyper}.
We choose the best hyper-parameter combination on the development set and test it on the evaluation server.

\subsubsection{Results}

\begin{figure*}[t]
\centering
\subfigure[MNLI.]{\includegraphics[width=0.23\textwidth]{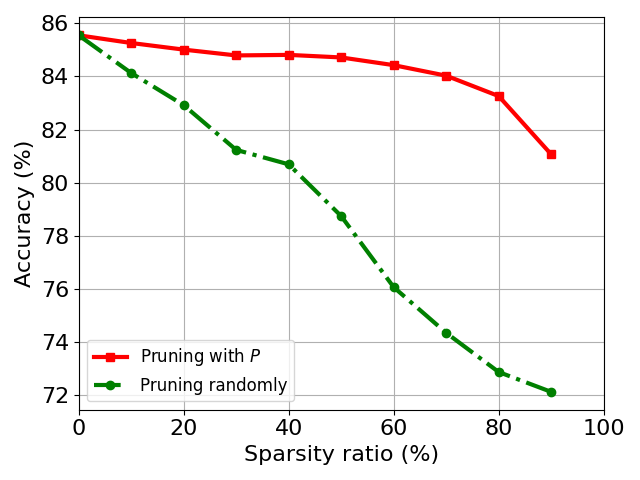}}
\subfigure[QQP.]{\includegraphics[width=0.23\textwidth]{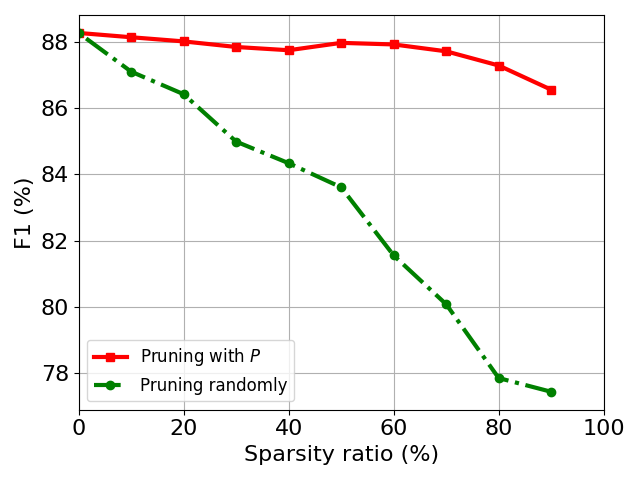}}
\subfigure[QNLI.]{\includegraphics[width=0.23\textwidth]{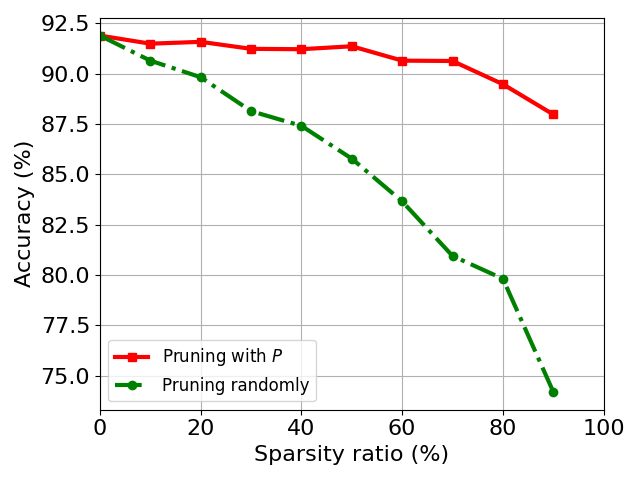}}
\subfigure[SST-2.]{\includegraphics[width=0.23\textwidth]{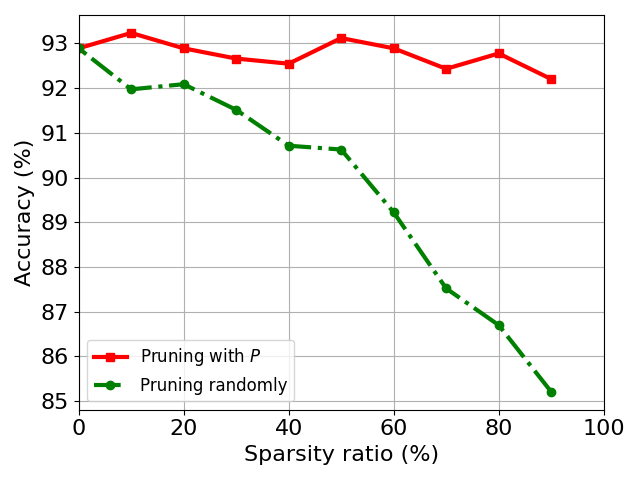}}
\subfigure[CoLA.]{\includegraphics[width=0.23\textwidth]{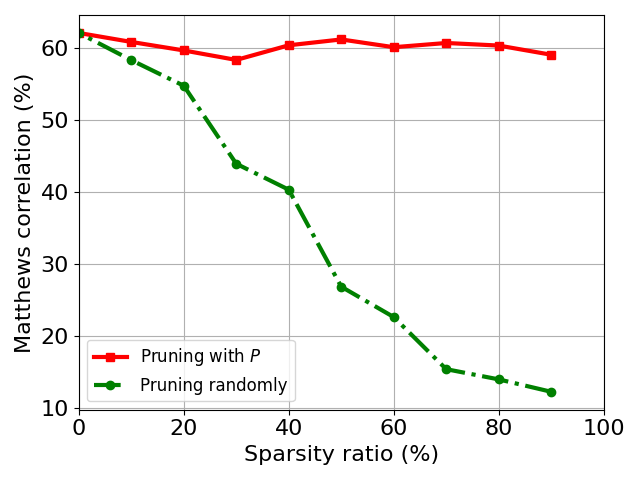}}
\subfigure[STS-B.]{\includegraphics[width=0.23\textwidth]{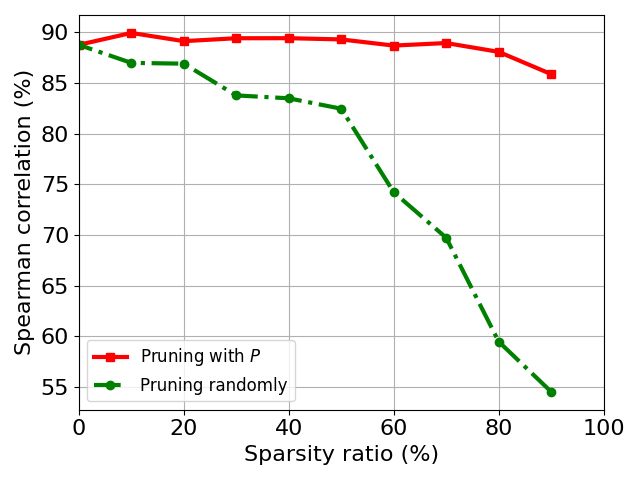}}
\subfigure[MRPC.]{\includegraphics[width=0.23\textwidth]{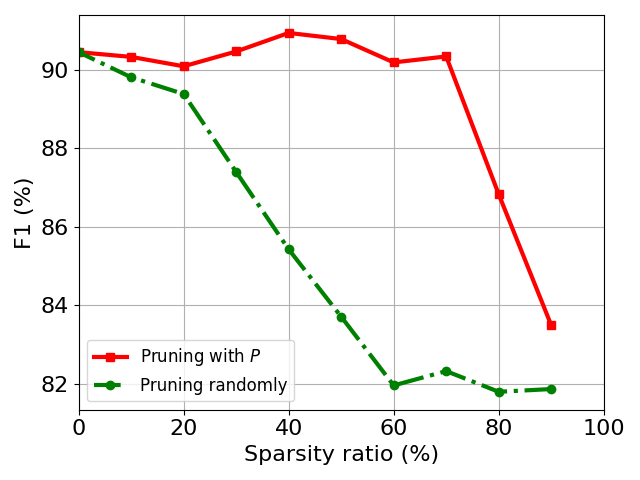}}
\subfigure[RTE.]{\includegraphics[width=0.23\textwidth]{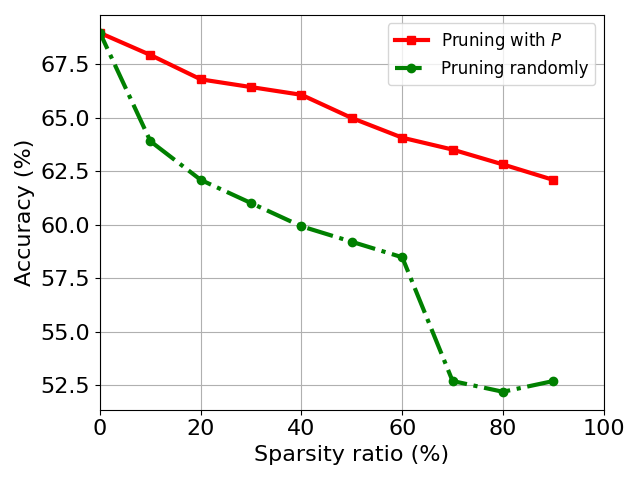}}
\caption{Performance of the BERT-base at different attention mask sparsity ratios on the GLUE development set.
MNLI is the average performance on the MNLI-m and MNLI-mm sections.}
\label{fig:iter_pruning}
\end{figure*}

Results on the GLUE benchmark are shown in
Table~\ref{tab:glue}.
For comparison, we also show the BERT-base results reported in
\citet{devlin2019bert}, and BERT-base (ours) is reproduced by ourselves.
As can be seen, even by
constraining the self-attention mechanism such that each token cannot attend to
itself, the performance of this constrained model
``BERT-base (no diag-attention)"
is still comparable
with the original
or even better
than BERT-base. On the other hand, when the
sparse attention masks are
randomly chosen,
the performance drops.
Table~\ref{tab:swag}
shows the results on SWAG and SQuAD, and
the observations
are similar.
This demonstrates that attentions to the diagonal positions are not necessary. Without
attending to self-token, the model performance is not deteriorated.

\vspace{-0.1em}
\subsection{Progressive Pruning of Self-attention\label{sec:pruning}}

To further investigate the effectiveness of the searched mask, we perform progressive pruning of self-attention according to the results in Section~\ref{sec:visual}.
Experiments are performed on the GLUE benchmark and the experiment settings are the same.
We
threshold
$\boldsymbol{P}$ to a binary attention mask $\boldsymbol{M}$ so that a part of entries in $\boldsymbol{P}$ are pruned.
We investigate the performance with different mask sparsity ratios.
Specifically,
for each head,
we remove the
smallest
$10\%/20\%/\dots/90\%$ entries
from $\boldsymbol{P}$ and keep other positions active.
For
comparison, we include a random baseline that randomly removes the same number of
entries.

As can be seen from Figure~\ref{fig:iter_pruning}, pruning by the attention distribution matrix $\boldsymbol{P}$
consistently outperforms random pruning on all GLUE tasks.
This verifies the importance analysis in Section~\ref{sec:visual}.
Moreover,
models from
different GLUE sub-tasks
can have different degrees of redundancy.
For instance, performance on CoLA sub-task
remains stable when sparsity ratio gets higher.
In contrast, the RTE sub-task performance shows a significant drop
as the model is
sparsified, illustrating the need for a denser self-attention computation.
Thus, we can select different sparsity ratios for different tasks to balance performance and efficiency.

\section{SparseBERT}

As discussed in Section \ref{subsec:Efficient-Transformer}, vanilla
self-attention suffers from quadratic complexity, and a number of
sparse Transformers
have been proposed
\cite{guo2019star,child2019generating,li2019enhancing,beltagy2020longformer,zaheer2020big}.
However, their
attention masks
are manually designed. In this section, we use the observations in the previous
sections to help
develop a series of sparse attention masks with different sparsity ratios.

\vspace{-2mm}
\subsection{Differentiable Attention Mask}

A straightforward method to generate sparse Transformers is by pruning the
attention distribution $\boldsymbol{P}$ as is performed
in Section~\ref{sec:pruning}.
However, this involves two separate stages. The resultant performance may be
sub-optimal as the whole model is not trained end-to-end. Moreover,
discretization of the continuous attention distribution $\boldsymbol{P}$ may mislead the final attention mask.

\begin{figure*}[b]
\centering
\subfigure[MNLI.]{\includegraphics[width=0.23\textwidth]{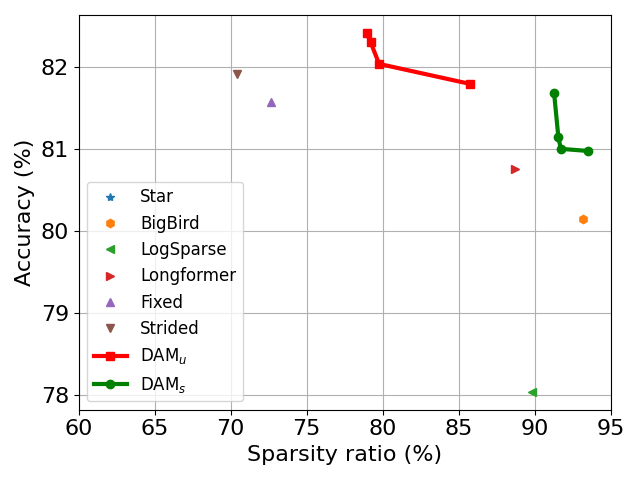}}
\subfigure[QQP.]{\includegraphics[width=0.23\textwidth]{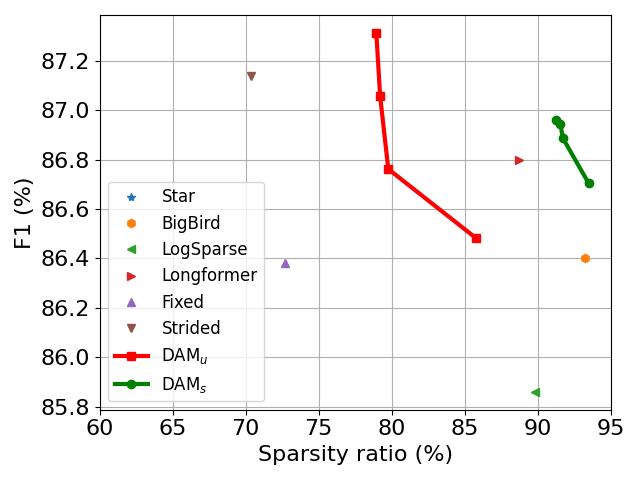}}
\subfigure[QNLI.]{\includegraphics[width=0.23\textwidth]{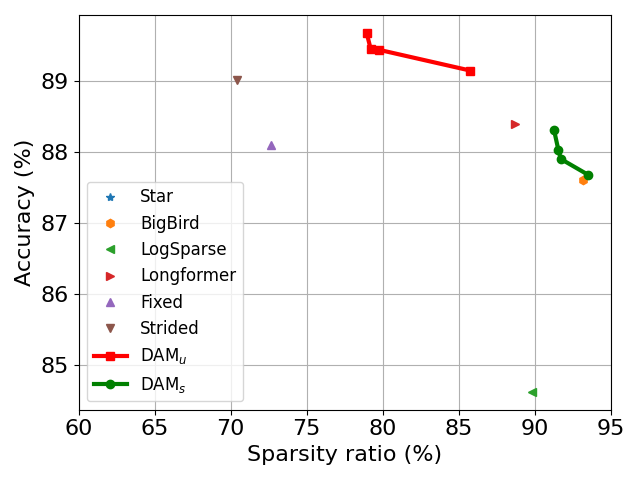}}
\subfigure[SST-2.]{\includegraphics[width=0.23\textwidth]{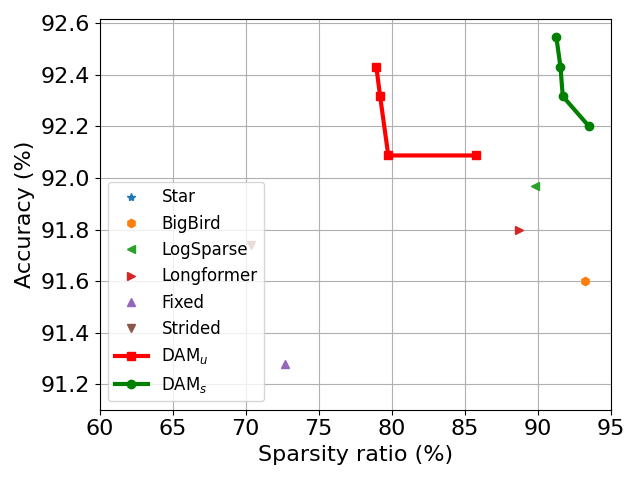}}
\subfigure[CoLA.]{\includegraphics[width=0.23\textwidth]{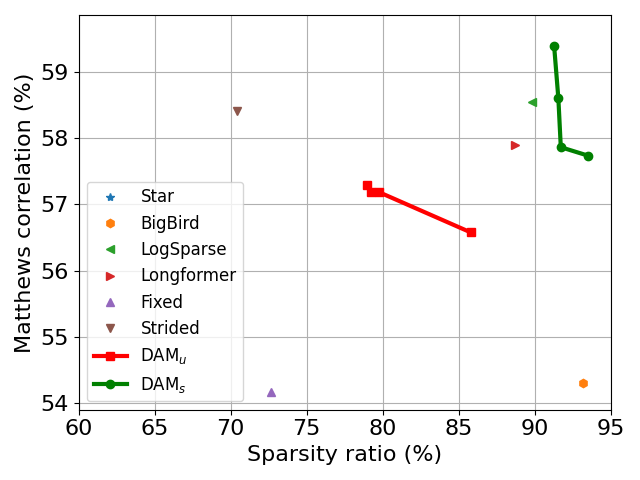}}
\subfigure[STS-B.]{\includegraphics[width=0.23\textwidth]{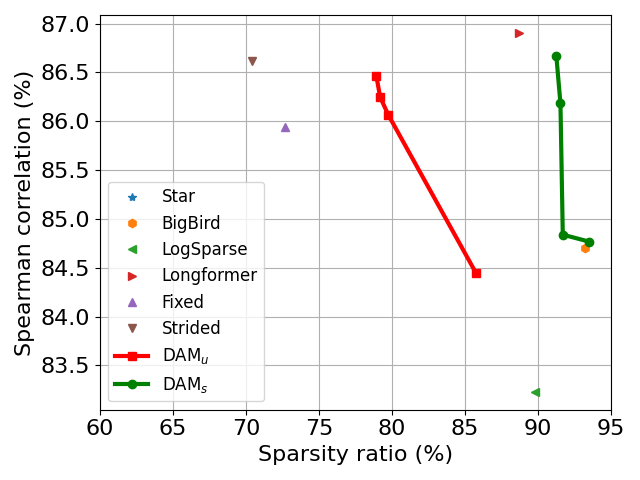}}
\subfigure[MRPC.]{\includegraphics[width=0.23\textwidth]{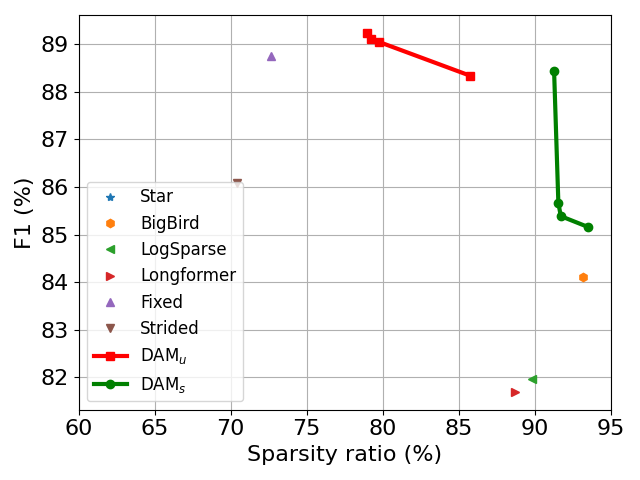}}
\subfigure[RTE.]{\includegraphics[width=0.23\textwidth]{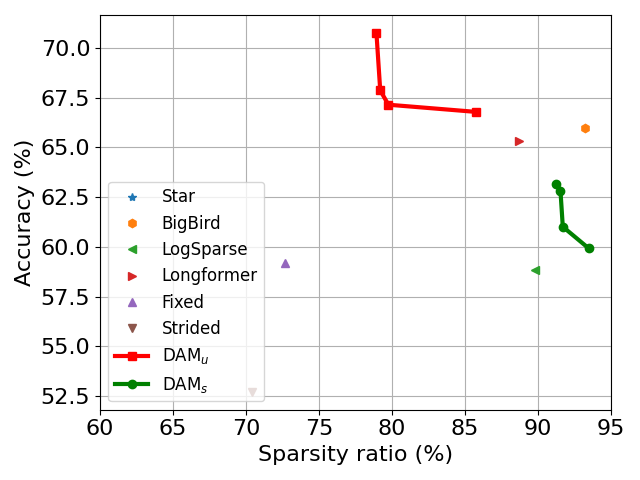}}
\caption{Performance of the BERT-base for different attention masks on the GLUE development set. MNLI shows the average performance on the MNLI-m and MNLI-mm sections.}
\label{fig:glue_mask}
\end{figure*}

To enable
end-to-end training, we propose to use the Gumbel
relaxation \cite{maddison2017concrete}.
Instead of using the sigmoid function
to output the attention probability
as
in (\ref{eq:P}),
we use the Gumbel-sigmoid:
{\small
\[
M_{i,j} =\text{Gumbel-sigmoid}(\alpha_{i,j}) = \text{sigmoid}((\alpha_{i,j}+G_1-G_2)/\tau),
\]
}where
$G_1$, $G_2$ are independent Gumbel noises generated from the
uniform distribution $U$ as:
{\small
\[ G_k =-\log(-\log(U_k)), U_k\sim U(0, 1), \]}and $\tau$
is a temperature hyperparameter.
As $\tau$ approaches zero, the
Gumbel-sigmoid output becomes a discrete distribution in $\{0,1\}$.
Thus, we can train the attention mask in an end-to-end manner with the
Gumbel-sigmoid variant. To balance mask sparsity  with performance,
we add the sum absolute values of the attention mask
to the loss, as:
\begin{equation} \label{eq:tradeoff}
\mathcal{L}=l(\BERT(\boldsymbol{X},\boldsymbol{A}(\boldsymbol{X})\odot \boldsymbol{M}(\boldsymbol{\alpha});\boldsymbol{w}))+\lambda\|\boldsymbol{M}(\boldsymbol{\alpha})\|_{1},
\end{equation}
where
$l(\BERT(\boldsymbol{X},\boldsymbol{A}(\boldsymbol{X});\boldsymbol{w}))$
is the pre-training loss, and
$\lambda$ is a trade-off hyperparameter.
When $\lambda$ is large,
we pay more emphasis on efficiency and the learned attention mask $\boldsymbol{M}$ is
sparser, and vice versa.
We also apply the renormalization trick to normalize the attention distributions in Eq. (\ref{eq:tradeoff}).
The obtained one-hot
Gumbel sigmoid output
can then be directly used as the attention mask.
The whole algorithm is shown in Algorithm~\ref{alg:DAM}.
We optimize both parameter sets simultaneously in one-level optimization until convergence and the generated mask is returned finally.

\begin{algorithm}[t]
\caption{Differentiable Attention Mask (DAM).\label{alg:DAM}}
\begin{algorithmic}[1]
\STATE initialize model parameter $\boldsymbol{w}$ and attention mask parameter $\boldsymbol{\alpha}$.
\REPEAT
\STATE generate mask $M_{i,j}\leftarrow \text{gumbel-sigmoid}(\alpha_{i,j})$;
\STATE obtain the loss with attention mask $\mathcal{L}$;
\STATE update parameter $\boldsymbol{w}$ and $\boldsymbol{\alpha}$ simultaneously;
\UNTIL{convergence.}
\STATE{\textbf{return} attention mask $\boldsymbol{M}$.}
\end{algorithmic}
\end{algorithm}

Note that the attention mask obtained in Algorithm~\ref{alg:DAM} is unstructured,
as the
$\alpha_{i,j}$'s are all independent of each other. This
irregular structure may affect the efficiency of the final CUDA implementation.
To alleviate this problem, we use the observed sparsity patterns in
Figure~\ref{fig:importance_sc} to constrain the structure of the attention mask.
First,
as the special tokens are important,
we require the first and last row/column
of the attention mask to be active.
Second,
for all positions on each line parallel to the diagonal, we share their mask
parameters including their two Gumbel noises
such that the
generated mask has $M_{i,j}=M_{i+k,j+k}$ for integer $k$.
Among the previously proposed
attention masks, the Sparse Transformer (strided) \cite{child2019generating} and
LogSparse Transformer \cite{li2019enhancing} conform to our definition of
structured attention masks, and they can be implemented efficiently by custom CUDA kernels.
With the constrained structure, there are now only $n-2$ attention mask parameters.
Therefore, the search space is reduced to $2^{n-2}$ in structured attention mask search.
Since the sparsity ratio directly affects the efficiency of the SparseBERT, we will
show the performance of the attention mask with its sparsity ratio.

\subsection{Experiments}
As in previous sections, we evaluate the proposed DAM
algorithm by using
the BERT-base \cite{devlin2019bert} model
on the
GLUE data sets \citep{wang2018glue} with pre-training and fine-tuning paradigm
\citep{devlin2019bert}.
We use the same pre-training settings as in
Section~\ref{subsec:Continuous-Relaxation} with the dynamic attention mask.
As for fine-tuning, the attention mask is fixed and other experiment settings are the same as Section~\ref{sec:glue}.
The DAM variant
using unstructured attention mask
is denoted
DAM$_u$,  while the one using
structured attention mask  is denoted
DAM$_s$.
The trade-off hyperparameter $\lambda$ in (\ref{eq:tradeoff})
is varied
in $\{10^{-1}, 10^{-2}, 10^{-3}, 10^{-4}\}$
to generate masks with different sparsities
for both structured and unstructured attention masks.

For comparison, we consider the following
baselines
with different manually-designed attention masks
(Figure~\ref{fig:The-visualization-of-1}):
(i) Star \citep{guo2019star}; (ii) LogSparse \citep{li2019enhancing}; (iii)
Strided
\citep{child2019generating};
(iv) Fixed
\citep{child2019generating}; (v) Longformer \cite{beltagy2020longformer}; and
(vi) BigBird
\cite{zaheer2020big}.
For the Longformer, we set the sliding window size to $2$ and
two input locations
are randomly selected for global attention.
For the BigBird, we ignore its block structure for fair comparison and set its random attention size to $2$, window
attention size to $1$, and global attention size to $2$.
The sparsity ratios of these fixed attention masks can be easily computed.

Performance on GLUE sub-tasks are shown in Figure~\ref{fig:glue_mask} and
Table~\ref{tab:glue_mask}.
As can be seen, attention masks generated by the proposed method outperform
manual-designed masks in almost all cases, and their sparsity ratios can be controlled by $\lambda$.
For example, compared with BigBird, the proposed DAM$_s$ ($\lambda=10^{-1}$)
achieves higher average performance while using a
sparser
attention mask.

\begin{table*}[t]
\caption{Comparison of BERT-base model with different attention masks on the GLUE development set (\%).\label{tab:glue_mask}}
\centering
\resizebox{0.95\textwidth}{!}{
\begin{tabular}{lcccccccccc}
\hline
 &Sparsity ratio& \multirow{1}{*}{MNLI-(m/mm)} & QQP & QNLI & SST-2 & COLA & STS-B & MRPC & RTE & Average\tabularnewline
\hline
Strided \citep{child2019generating}    & 70.4 & 81.9/81.9 & 87.1 & 89.0 & 91.7 & 58.4 & 86.6 & 86.1 & 52.7 & 79.5 \tabularnewline
Fixed  \citep{child2019generating}    & 72.7 & 81.4/81.8 & 86.4 & 88.1 & 91.3 & 54.2 & 85.9 & 88.7 & 59.2 & 79.7 \tabularnewline
Longformer \citep{beltagy2020longformer} & 88.7 & 80.5/81.0 & 86.8 & 88.4 & 91.8 & 57.9 & 86.9 & 81.7 & 65.3 & 80.1 \tabularnewline
LogSparse \citep{li2019enhancing} & 89.8 & 77.9/78.2 & 85.9 & 84.6 & 92.0 & 58.5 & 83.2 & 82.0 & 58.8 & 77.9 \tabularnewline
BigBird \citep{zaheer2020big}   & 93.2 & 80.2/80.1 & 86.4 & 87.6 & 91.6 & 54.3 & 84.7 & 84.1 & 66.0 & 79.4 \tabularnewline
Star \citep{guo2019star}   & 96.1 & 79.1/79.0 & 86.2 & 86.4 & 91.2 & 59.6 & 84.7 & 83.9 & 60.3 & 78.9 \tabularnewline \hline
DAM$_u$ ($\lambda=10^{-4}$)    & 78.9 & 82.2/82.6 & 87.3 & 89.7 & 92.4 & 57.3 & 86.5 & 89.2 & 70.8 & 82.0 \tabularnewline
DAM$_u$ ($\lambda=10^{-3}$)    & 79.2 & 82.2/82.4 & 87.1 & 89.5 & 92.3 & 57.2 & 86.2 & 89.1 & 67.9 & 81.6 \tabularnewline
DAM$_u$ ($\lambda=10^{-2}$)    & 79.8 & 81.7/82.3 & 86.8 & 89.4 & 92.1 & 57.2 & 86.1 & 89.0 & 67.1 & 81.3 \tabularnewline
DAM$_u$ ($\lambda=10^{-1}$)    & 85.8 & 81.4/82.2 & 86.5 & 89.1 & 92.1 & 56.6 & 84.4 & 88.3 & 66.8 & 80.8 \tabularnewline \hline
DAM$_s$ ($\lambda=10^{-4}$)  & 91.2 & 81.7/81.7 & 87.0 & 88.3 & 92.5 & 59.4 & 86.7 & 88.4 & 63.2 & 80.9 \tabularnewline
DAM$_s$ ($\lambda=10^{-3}$)   & 91.6 & 81.0/81.2 & 86.9 & 88.0 & 92.4 & 58.6 & 86.2 & 85.7 & 62.8 & 80.3 \tabularnewline
DAM$_s$ ($\lambda=10^{-2}$)   & 91.7 & 81.1/80.9 & 86.9 & 87.9 & 92.3 & 57.9 & 84.8 & 85.4 & 61.0 & 79.8 \tabularnewline
DAM$_s$ ($\lambda=10^{-1}$)   & 93.5 & 80.9/81.0 & 86.7 & 87.7 & 92.2 & 57.7 & 84.8 & 85.2 & 59.9 & 79.6 \tabularnewline
\hline
\end{tabular}}
\end{table*}

\begin{figure*}[t]
\begin{centering}
\subfigure[Head-1.]{\includegraphics[width=0.16\textwidth]{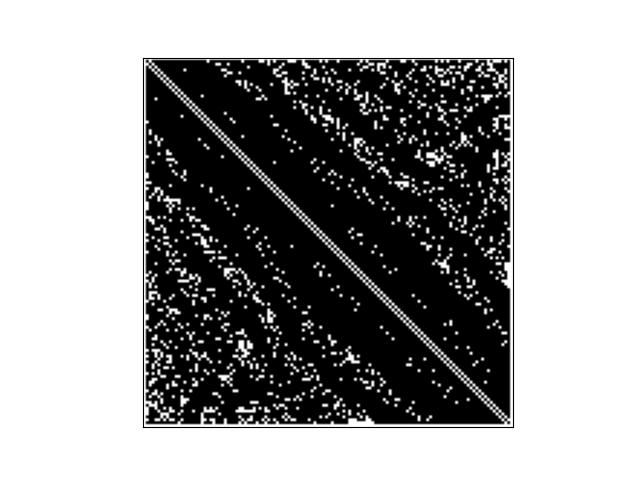}}
\subfigure[Head-2.]{\includegraphics[width=0.16\textwidth]{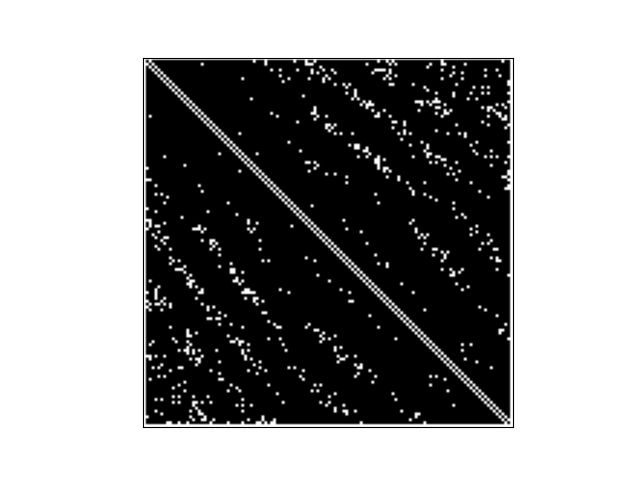}}
\subfigure[Head-3.]{\includegraphics[width=0.16\textwidth]{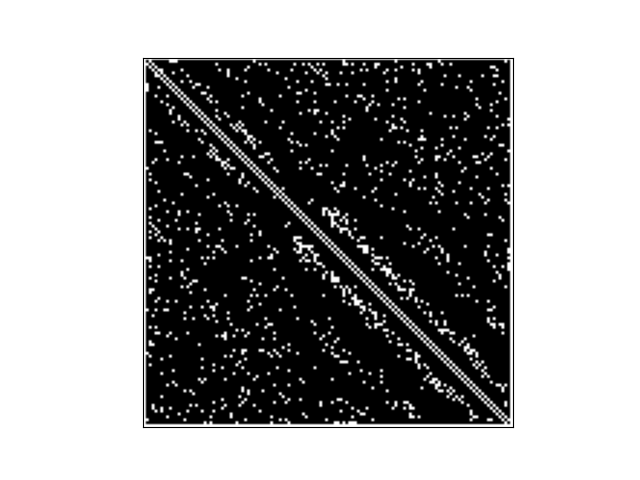}}
\subfigure[Head-4.]{\includegraphics[width=0.16\textwidth]{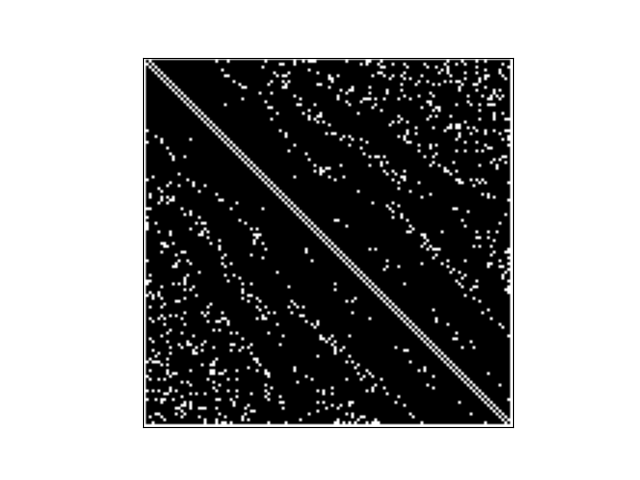}}
\subfigure[Head-5.]{\includegraphics[width=0.16\textwidth]{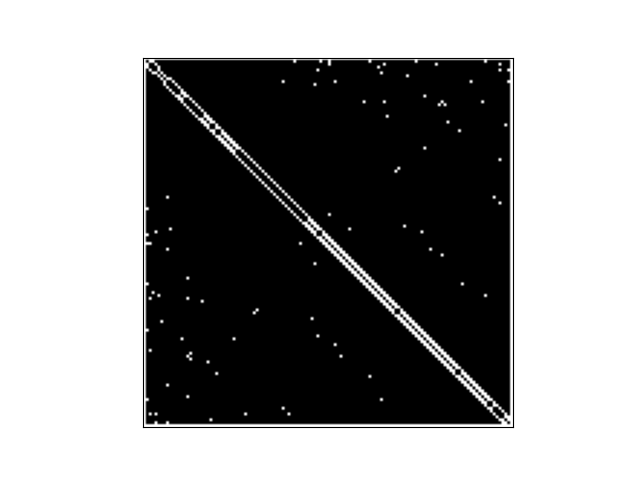}}
\subfigure[Head-6.]{\includegraphics[width=0.16\textwidth]{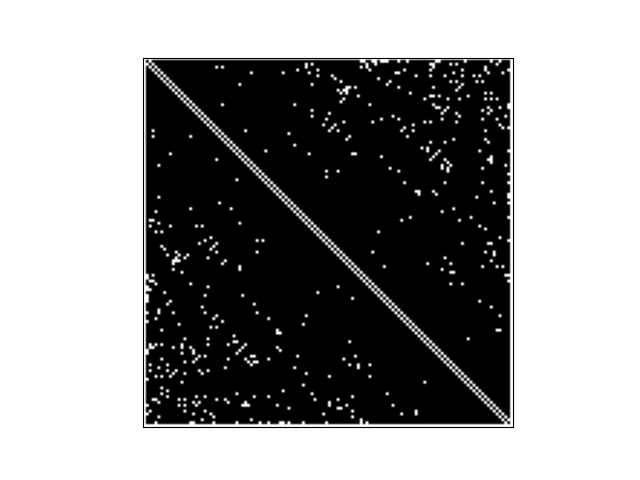}}
\subfigure[Head-7.]{\includegraphics[width=0.16\textwidth]{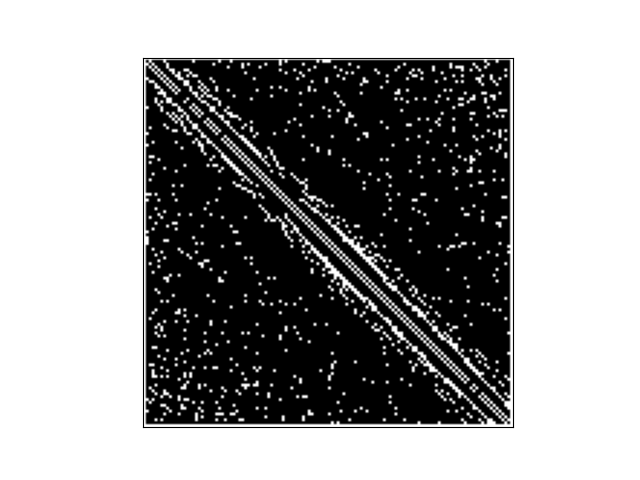}}
\subfigure[Head-8.]{\includegraphics[width=0.16\textwidth]{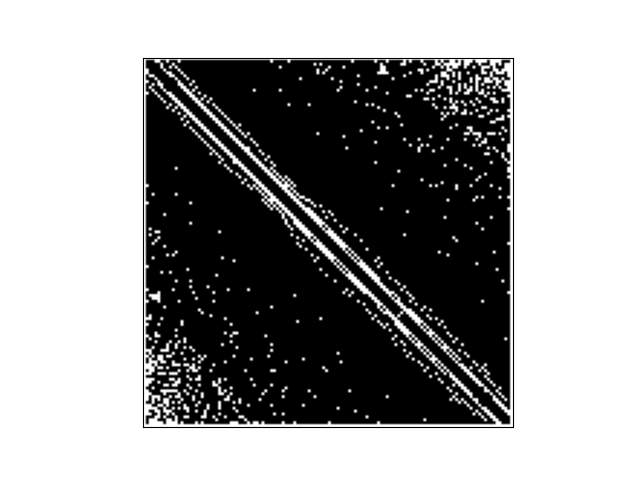}}
\subfigure[Head-9.]{\includegraphics[width=0.16\textwidth]{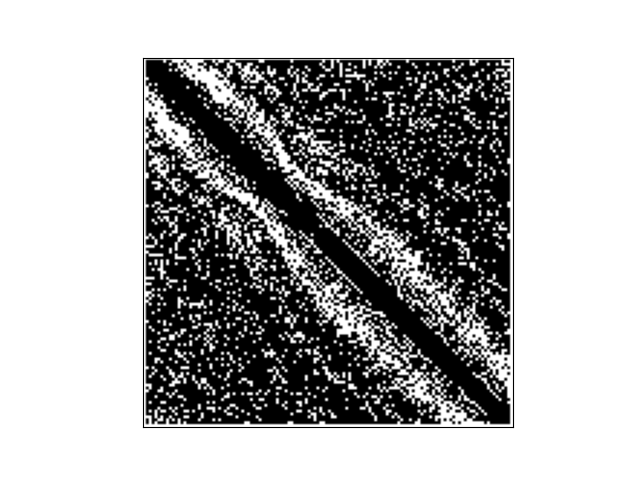}}
\subfigure[Head-10.]{\includegraphics[width=0.16\textwidth]{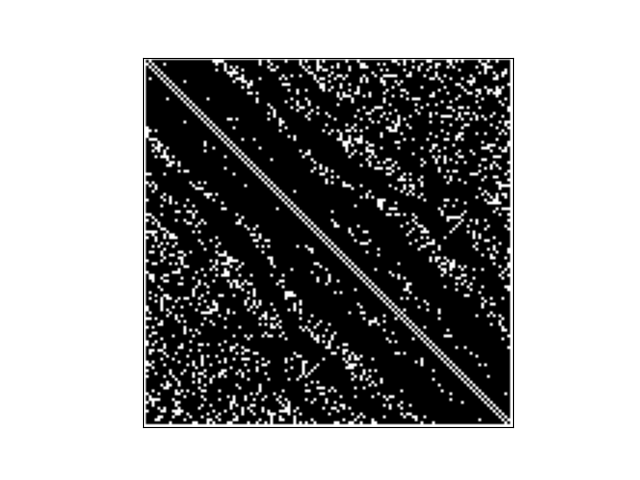}}
\subfigure[Head-11.]{\includegraphics[width=0.16\textwidth]{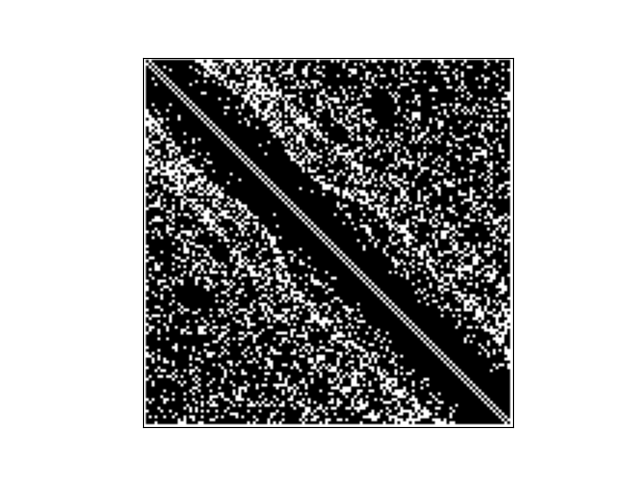}}
\subfigure[Head-12.]{\includegraphics[width=0.16\textwidth]{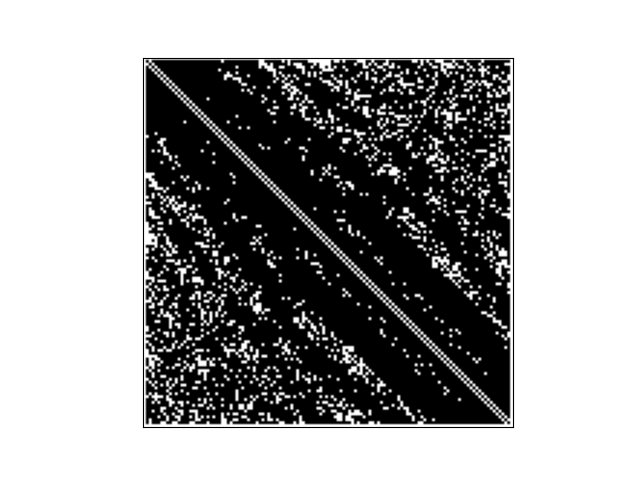}}
\par\end{centering}
\caption{Visualization of attention masks generated by DAM$_u$ ($\lambda=10^{-1}$). White means
with-attention and dark means no-attention. \label{fig:dam_u}}
\end{figure*}
\begin{figure*}[t]
\begin{centering}
\subfigure[Head-1.]{\includegraphics[width=0.16\textwidth]{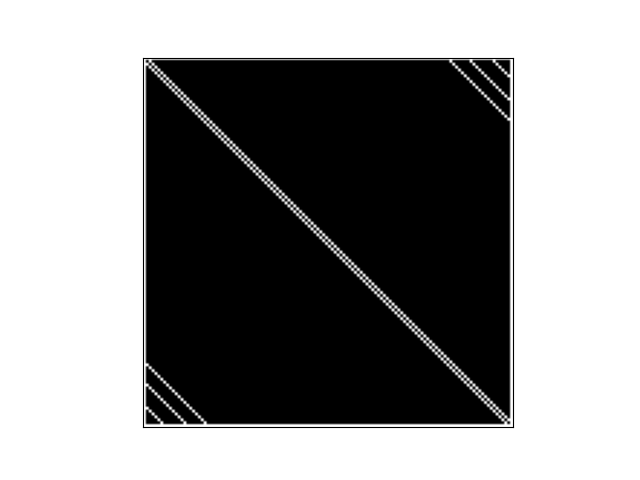}}
\subfigure[Head-2.]{\includegraphics[width=0.16\textwidth]{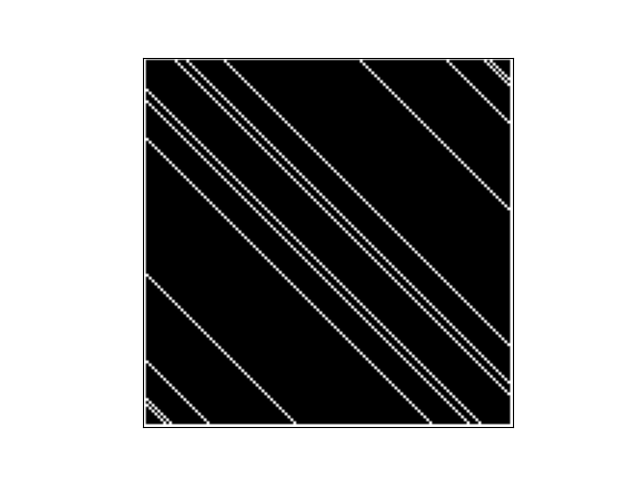}}
\subfigure[Head-3.]{\includegraphics[width=0.16\textwidth]{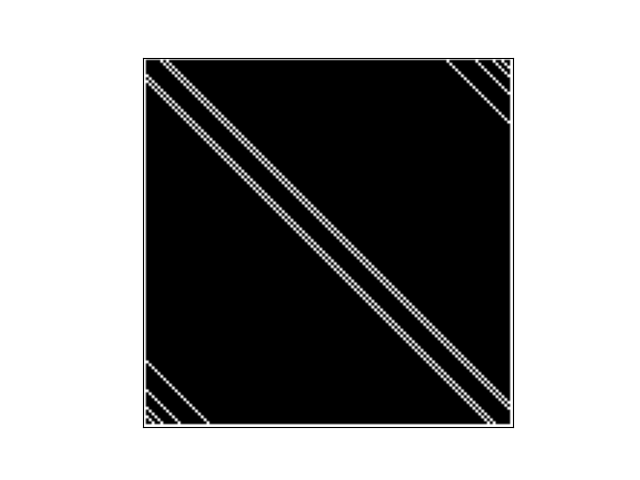}}
\subfigure[Head-4.]{\includegraphics[width=0.16\textwidth]{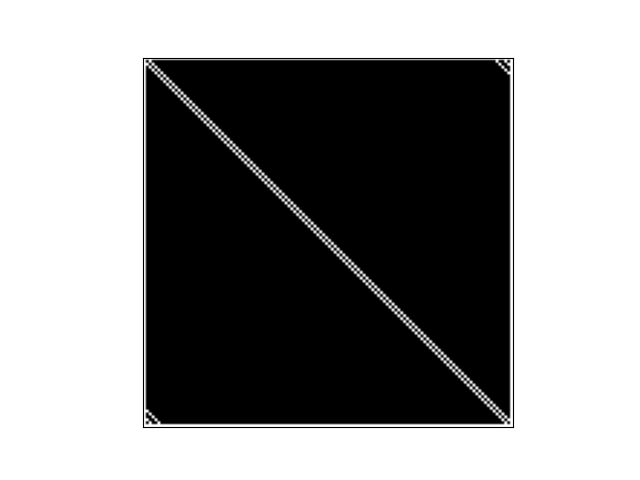}}
\subfigure[Head-5.]{\includegraphics[width=0.16\textwidth]{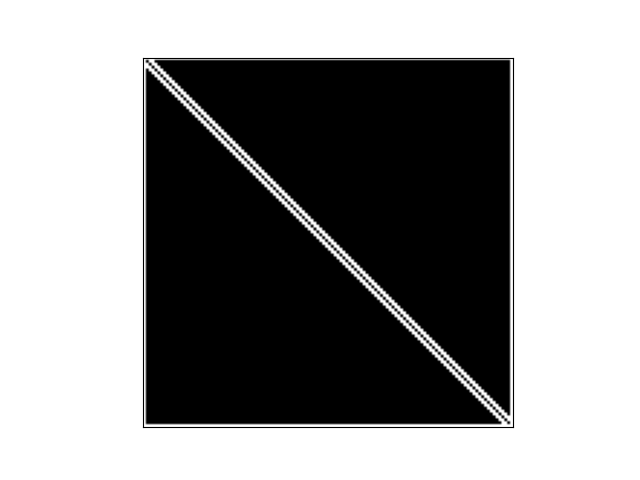}}
\subfigure[Head-6.]{\includegraphics[width=0.16\textwidth]{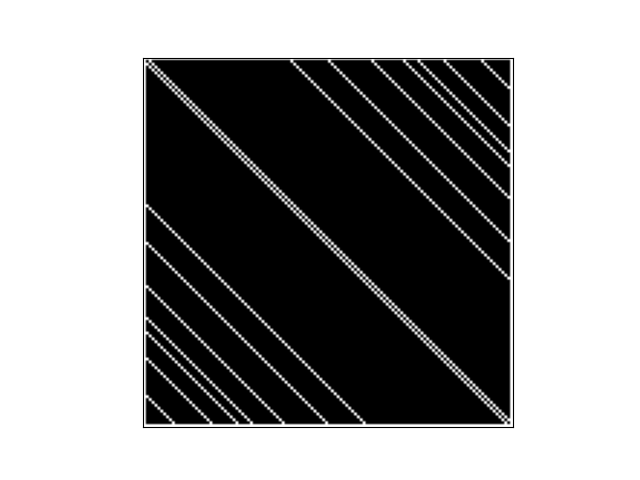}}
\subfigure[Head-7.]{\includegraphics[width=0.16\textwidth]{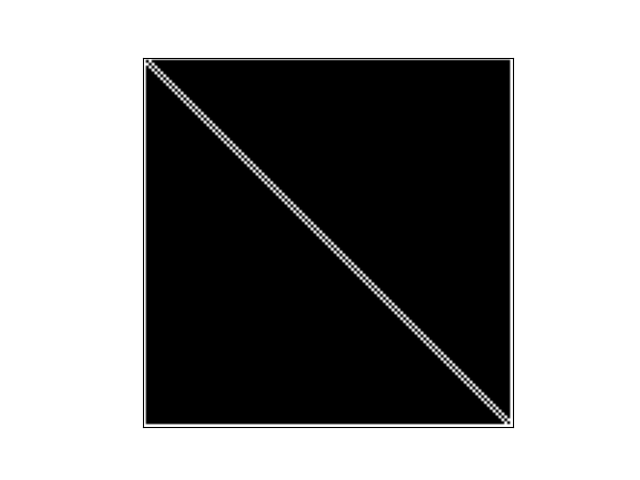}}
\subfigure[Head-8.]{\includegraphics[width=0.16\textwidth]{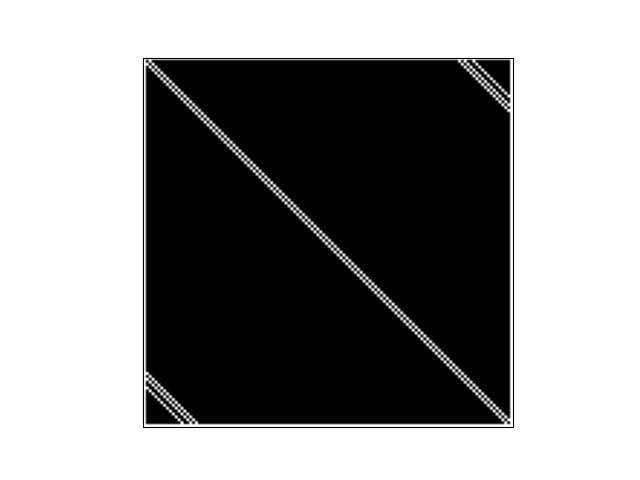}}
\subfigure[Head-9.]{\includegraphics[width=0.16\textwidth]{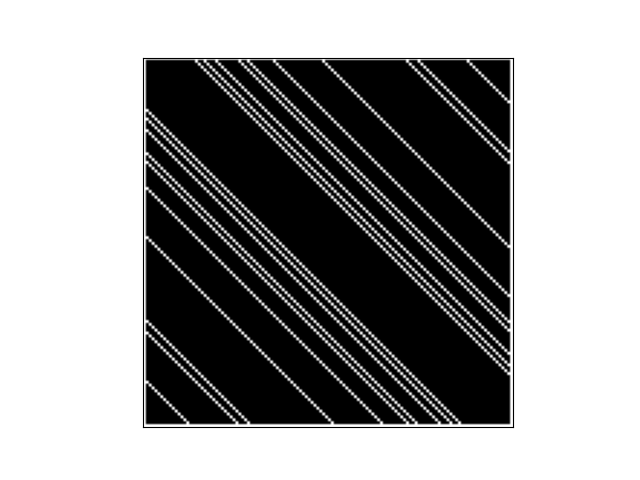}}
\subfigure[Head-10.]{\includegraphics[width=0.16\textwidth]{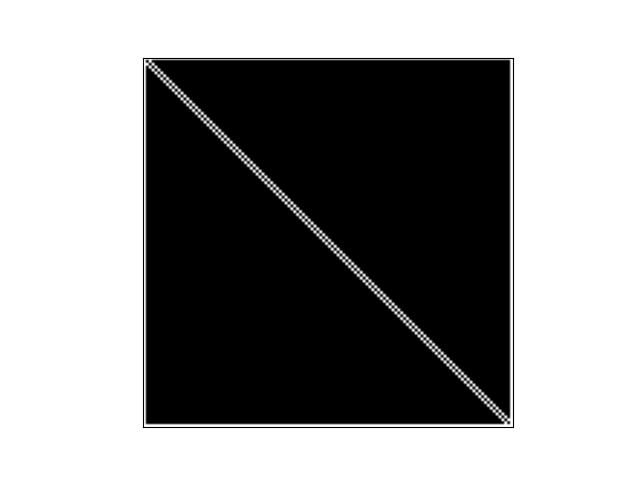}}
\subfigure[Head-11.]{\includegraphics[width=0.16\textwidth]{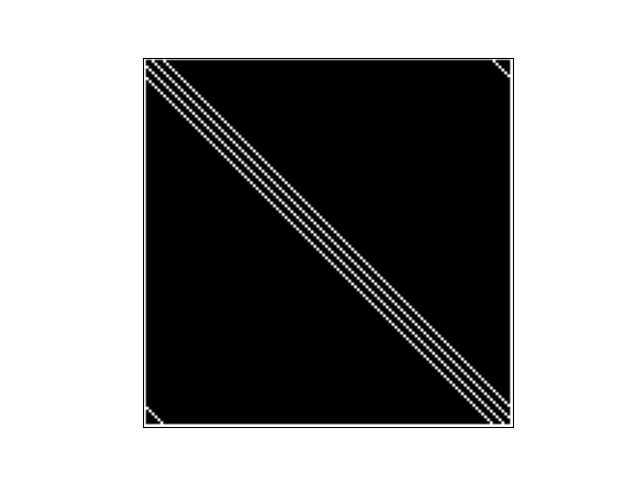}}
\subfigure[Head-12.]{\includegraphics[width=0.16\textwidth]{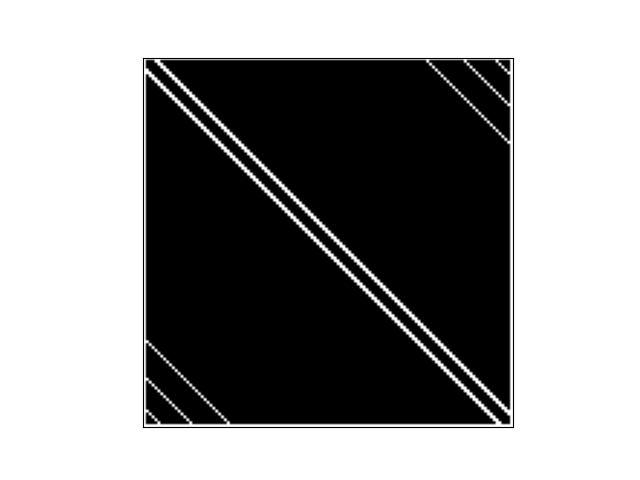}}
\par\end{centering}
\caption{Visualization of attention masks generated by DAM$_s$ ($\lambda=10^{-1}$).  White means
with-attention and dark means no-attention. \label{fig:dam_s}}
\end{figure*}

Figures~\ref{fig:dam_u} and \ref{fig:dam_s} show two ensembles of attention masks
generated by DAM$_u$ and DAM$_s$, respectively, after pre-training with $\lambda=10^{-1}$.
As can be seen, the diagonal
entries are not attended, while their neighborhood positions are attended in most heads.
From the visualizations of attention masks in
Figure~\ref{fig:The-visualization-of-1} and Figures~\ref{fig:dam_u}-\ref{fig:dam_s}, we can summarize
them into three categories: (i) Stride/Fixed/Logsparse, which only contain
neighboring attention; (ii) BigBird/Longformer/Star, which contain both
neighboring attention and attention from special tokens; (iii) The proposed
attention masks, which replace diag-attention with other attention positions from the second category.
From the empirical performance on the GLUE benchmark, the first class is the worst, the second
is competing while ours are the best, which agrees with the observations in Section~\ref{sec:visual}. Moreover,  DAM learns different attention masks for different heads, which utilizes the multi-head structure of the model.

\subsection{Ablation Study}
\subsubsection{Effect of Diag-attention for Masks}
In Section~\ref{sec:Which-attention-position}, we show the unimportance of
diag-attention in the self-attention module. Here, we study the effect of
diag-attention in different attention masks. Specifically, we perform ablation experiments for different attention masks and compare their average scores on the GLUE development set.
We illustrate the results on the existing attention masks
(Figure~\ref{fig:The-visualization-of-1}) and two structured masks of SparseBERT in Table~\ref{tab:rebuttal}.

As can be seen, the average GLUE score of using diag-attention and dropping diag-attention is similar for all attention masks. Thus, dropping diag-attention can increase the sparsity ratio further without harming performance.

\subsubsection{One-stage vs Two-stage Pruning}
In this section, we compare
Differentiable Attention Mask described in Algorithm~\ref{alg:DAM}, which
generates
the attention mask as part of the end-to-end training process
(one-stage) with the pruning approach in
Section~\ref{sec:pruning}, which
first obtains the attention probabilities in $\boldsymbol{P}$ and then performs thresholding to
obtain the binary attention mask
(two-stage).
For the two-stage attention mask, we prune $80\%/85\%/90\%/95\%$ entries of self-attention for better comparison.

\begin{figure}[t]
\begin{centering}
\subfigure[QNLI.]{\includegraphics[width=0.24\textwidth]{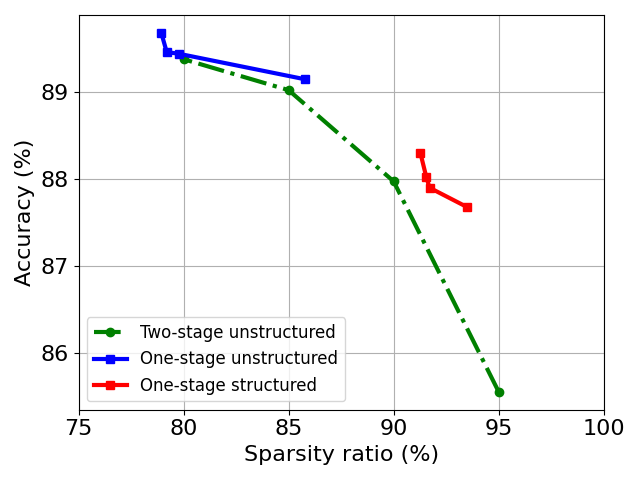}}
\subfigure[MRPC.]{\includegraphics[width=0.236\textwidth]{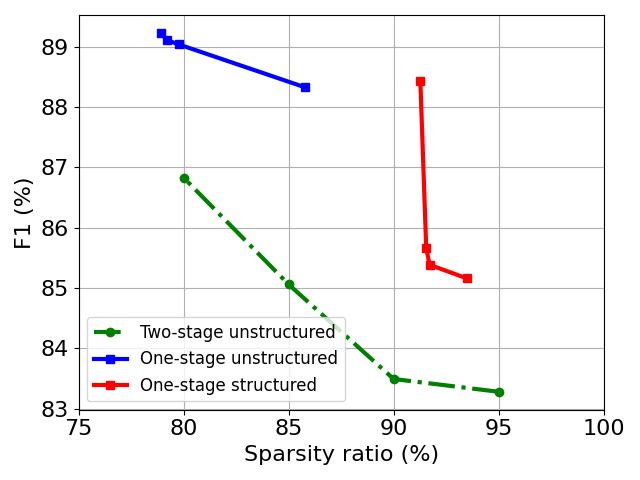}}
\par\end{centering}
\caption{Comparison of one-stage and two-stage attention mask on QNLI and MRPC development set. \label{fig:ablation}}
\end{figure}

\begin{table*}[t]
    \centering
    \caption{Ablation study on the importance of diag-attention in different
	 attention masks. Here, ``w/" means using diag-attention and ``w/o" means
	 without using diag-attention. As can be seen, dropping diag-attention
	 increases sparsity ratio without harming the performance.}
    \resizebox{2\columnwidth}{!}{
    \begin{tabular}{ccc|cc|cc|cc|cc|cc|cc|cc}
    \hline
    & \multicolumn{2}{c|}{Strided} & \multicolumn{2}{c|}{Fixed} &
	 \multicolumn{2}{c|}{Longformer} & \multicolumn{2}{c|}{LogSparse} &
	 \multicolumn{2}{c|}{BigBird} & \multicolumn{2}{c|}{Star}  &
	 \multicolumn{2}{c|}{DAM$_s$($\lambda=10^{-4}$)} & \multicolumn{2}{c}{DAM$_s$($\lambda=10^{-1}$)}\tabularnewline
    & w/ & w/o & w/ & w/o & w/ & w/o & w/ & w/o & w/ & w/o & w/ & w/o  & w/ & w/o  & w/ & w/o
	 \\ \hline
Sparsity (\%) & 70.4 & 71.2 & 72.7 & 73.4 & 88.7 & 89.5 & 89.8 & 90.6 & 93.2 & 93.9 & 96.1 & 96.9 & 90.4 & 91.2 & 92.7 & 93.5\tabularnewline
GLUE (\%) & 79.5 & 80.2 & 79.7 & 79.6 & 80.1 & 80.1 & 77.9 & 77.8 & 79.4 & 79.5 & 78.9 & 78.6 & 80.5 & 80.9 & 79.3 & 79.6\tabularnewline
\hline
\label{tab:rebuttal}
\end{tabular}}
\end{table*}

Here we only illustrate the performance comparison
on QNLI and MRPC development sets in Figure~\ref{fig:ablation}.
As can be seen, the attention masks
(unstructured and structured) generated by one-stage optimization achieve better
performance, which is caused by the gap between continuous $\boldsymbol{P}$ and discrete $\boldsymbol{M}$.
However,
the two-stage attention mask can easily adjust
its sparsity to any desired value,
while the one-stage approach
cannot set the sparsity
directly as it is controlled by the hyper-parameter $\lambda$
in Eq. (\ref{eq:tradeoff}).

\section{Conclusion}
In this paper, we investigate the importance of the different attention positions
in the self-attention mechanism. By
jointly optimizing
a soft attention mask
with the BERT model, we obtain several interesting findings. In
particular, one surprising observation is that the diagonal elements in the
attention matrix are the least important, which conflicts with observations in prior works. We then show,
both theoretically and experimentally, that
these diagonal elements are indeed not useful for universal approximation and
empirical performance.
Besides,
by using the Gumbel-sigmoid function,
we propose
to optimize the attention mask
in an end-to-end manner
for efficient Transformer design. Extensive experimental results on a number of NLP tasks
demonstrate the usefulness of the proposed algorithm.

\clearpage{}
\bibliography{example_paper}
\bibliographystyle{icml2021}

\clearpage{}

\appendix
\twocolumn[ \icmltitle{Appendix for \\ SparseBERT: Rethinking the Importance Analysis in Self-attention}
\vskip 0.3in ]
\section{Proof \label{app:proof}}
\subsection{Proof for Step 1}
\begin{lemma}[Lemma 8 \citep{yun2019Transformers}]
For any $f\in \mathcal{F}_{CD}$, there exists a piece-wise constant function $\bar{f}$ such that $d_p(f,\bar{f})<\epsilon /3$.
\end{lemma}
\begin{proof}
$f$ is uniformly continously since $f$ is a continous function on $[0,1]^{n\times d}$, which implies: \\
$\forall \epsilon >0, \exists\ \delta>0$, such that
\[
  \forall \boldsymbol{X}, \boldsymbol{Y}, ||\boldsymbol{X}-\boldsymbol{Y}||_{\infty}<\delta \Rightarrow ||f(\boldsymbol{X})-f(\boldsymbol{Y})||_p<\epsilon/3.
\]
Then we split the compact domain $[0,1]^{n\times d}$ into a grid of granularity $\delta$, such that $G_\delta\in\{0,\delta,\dots, 1\}$. By defining the following piece-wise constant function
\[
\bar{f}(\boldsymbol{X})=\sum_{\boldsymbol{G}\in G_\delta}f(\boldsymbol{G})*\mathbbm{1}\{\boldsymbol{X}\in \boldsymbol{G}+[0,\delta]^{n\times d}\},
\]
we have
\[
||f(\boldsymbol{X})-\bar{f}(\boldsymbol{X})||_p=||f(\boldsymbol{X})-f(\boldsymbol{G})||_p<\epsilon / 3.
\]
Thus,
\[
d_p(f,\bar{f})=(\int||f(\boldsymbol{X})-\bar{f}(\boldsymbol{X})||_p^pd\boldsymbol{X})^{1/p}<\epsilon /3.
\]
This proves the lemma.
\end{proof}

\subsection{Proof for Step 2}
\subsubsection{Quantization (Feed-forward)}
\begin{lemma}[Lemma 5 \citep{yun2019Transformers}]
  Consider a quantization mapping $g_q^{ent}$:
\begin{align*}
  &g_q^{ent}(t)\\
 = &\begin{cases}
k\delta& \text{if}\;k\delta\leq t<(k+1)\delta,k\in[1:1/\delta-1],\\
-\delta^{-nd}& \text{otherwise.}
\end{cases}
\end{align*}
There exists a function $g_q$ composed of $d/\delta+d$ token-wise feed-forward layers with $r=1$ and piece-wise linear functions (at most three pieces), such that the quantization is performed on each entry of the input.
\end{lemma}
We first quantize the input $\boldsymbol{X}$ to their corresponding grid $G_\delta$ by quantization function $g_q$.
\subsubsection{Contexual Mapping (Self-attention)}
This is the main difference between ours and previous works, where self-attention without diag-attention adds additional constraints to the attention matrix. We will illustrate the definition of contextual mapping first and then prove Transformer blocks without diag-attention can also reach contextual mapping.
\begin{definition} \label{def:1}
  (Contextual Mapping) For a set $G_\delta\in\mathbb{R}^{n\times d}$, a contextual mapping is a function mapping $q:G_\delta\rightarrow\mathbb{R}^n$ satisfying:
  \begin{itemize}
    \item For any $\boldsymbol{G}\in G_\delta$, all entries in $q(\boldsymbol{G})$ are  distinct.
    \item For any $\boldsymbol{G}_1, \boldsymbol{G}_2 \in G_\delta$ ($\boldsymbol{G}_1\neq \boldsymbol{G}_2$), all entries of $q(\boldsymbol{G}_1)$ and $q(\boldsymbol{G}_2)$ are distinct.
  \end{itemize}
\end{definition}
\begin{lemma}
  There exists a function $g_c$ composed of $\delta^{-d}+1$ self-attention layers without diag-attention, such that $q(\boldsymbol{G}):=g_c(\boldsymbol{G})u$ satisfies the contextual mapping definition.
\end{lemma}
\begin{proof}
  Consider the function $\psi$, which can be implemented by a self-attention without diag-attention:
\begin{align*}
  \psi(\boldsymbol{Z};b)_i&=\sigma_H[(\boldsymbol{Z}_{i,:}\boldsymbol{u}-b)(\boldsymbol{Z}_{j\neq i,:}\boldsymbol{u})^\top]\boldsymbol{Z}_{j\neq i,:}\boldsymbol{ue}^{(1)\top} \\
  &=
  \begin{cases}
    (\max_{j\neq i}\boldsymbol{Z}_{j,:}\boldsymbol{u})\boldsymbol{e}^{(1)\top} & \text{if}\;\boldsymbol{Z}_{i,:}\boldsymbol{u}>b,\\
    (\min_{j\neq i}\boldsymbol{Z}_{j,:}\boldsymbol{u})\boldsymbol{e}^{(1)\top} & \text{if}\;\boldsymbol{Z}_{i,:}\boldsymbol{u}<b,
  \end{cases}
\end{align*}
where $\boldsymbol{e}^{(1)}=[1,0,\dots,0]\in \mathbb{R}^d$ and $\boldsymbol{u}\in\mathbb{R}^d$ is an auxiliary vector, which will be selected later.

We can contrust a self-attention layer without diag-attention that consists of two such heads $\Psi(\boldsymbol{Z};b_1,b_2)= \psi(\boldsymbol{Z};b_1)-\psi(\boldsymbol{Z};b_2)$,
such that
\begin{align*}
&\Psi(\boldsymbol{Z};b_{1},b_{2})_{i,1} \\
& =
\begin{cases}
\max_{j\neq i}\boldsymbol{Z}_{j,:}\boldsymbol{u}\text{\textminus}\min_{j\neq i}\boldsymbol{Z}_{j,:}\boldsymbol{u}
& \text{if}\;b_{1}<\boldsymbol{Z}_{i,:}\boldsymbol{u}<b_{2}, \\
0
& \text{otherwise.}
\end{cases}
\end{align*}
Thus, if we define a self-attention layer without diag-attention of the form $\boldsymbol{Z}\rightarrow \boldsymbol{Z}+\delta^{-d}\Psi(\boldsymbol{Z};b_1,b_2)$,
then selective shift operation is performed.

Next, we select $\boldsymbol{u}=(1,\delta^{-1},\delta^{-2},\dots,\delta^{-d+1})$ and the following holds:
\begin{itemize}
  \item If $Z_{i,j}\neq -\delta^{-nd}$ for all $j$, then $\boldsymbol{Z}_{i,:}\boldsymbol{u}\in [0:\delta: \delta^{-d+1}-\delta]$. And the mapping from $\boldsymbol{Z}\in \{0,\delta,\dots,1-\delta\}^d$ to $[0:\delta:\delta^{-d+1}-\delta]$ is a bijective mapping.
  \item If $Z_{i,j}= -\delta^{-nd}$ for some $j$, then $\boldsymbol{Z}_{i,:}\boldsymbol{u}\leq-\delta^{-nd}+\delta^{-d+1}-\delta<0$.
\end{itemize}
Thus, the mapping $\boldsymbol{Z}_{i,:}\rightarrow \boldsymbol{Z}_{i,:}\boldsymbol{u}$ is a bijective mapping for $\{0,\delta,\dots,1-\delta\}^d$. We define $l_i=\boldsymbol{Z}_{i,:}\boldsymbol{u}$ and assume $l_1<l_2<\dots<l_n$ without loss of generality.

For each $l\in [0:\delta:\delta^{-d+1}-\delta]$, we choose $b_1= l-\delta/2, b_2= l+\delta/2$ and use $\delta^{-d}$ self-atteniton layers without diag-attention.
Only one row will be in the range $(b_1, b_2)$ each time and no other row will be affected. After above operation, $l_i$ becomes $\widetilde{l}_i$ for better clarification.

For $n$ rows, there are total $n$ phases for column updating. After each $i$ phases, we will maintain the following ordering:
\[
 l_{i+1}<l_{i+2}<\dots<l_n<\widetilde{l_1}<\widetilde{l_2}<\dots<\widetilde{l_i}.
\]

\textbf{Base Step} When $i=0$, it's the trivial case as
\[
l_1<l_2<\dots<l_n.
\]
When $i=1$, we have $\max_{j\neq 1}l_j=l_n$ and $\min_{j\neq 1}l_j=l_2$. $\widetilde{l}_1=\delta^{-d}(l_{n}-l_{2})+l_1$.
\begin{align*}
  \widetilde{l}_1-l_{n}
  &=\delta^{-d}(l_{n}-l_{2})+(l_1-l_{n})\\
  &>\delta^{-d}(\delta)-(\delta^{-d+1}-\delta) \\
  &=\delta^{-d+1}-\delta^{-d+1}+\delta \\
  &=\delta>0.
\end{align*}

\textbf{Inductive Step} When $1<i<n$, we have $\max_{j\neq i}l_j=\widetilde{l}_{i-1}$ and $\min_{j\neq i}l_j=l_{i+1}$. Thus, $\widetilde{l}_i=\delta^{-d}(\widetilde{l}_{i-1}-l_{i+1})+l_i$. By expansion, we have:
\begin{align*}
    \widetilde{l}_i
    & = (l_n-l_2)\delta^{-id} + \sum_{j=1}^{i-1}(l_j-l_{j+2})\delta^{-(i-j)d}+l_i.
\end{align*}
\begin{align*}
  &\widetilde{l}_i-\widetilde{l}_{i-1}=(l_n-l_2)(\delta^{-id}-\delta^{-(i-1)d})  \\
  &+\sum_{j=1}^{i-2}(l_j-l_{j+2})(\delta^{-(i-j)d}-\delta^{-(i-j-1)d}) \\
  &+ \delta^{-d}(l_{i-1}-l_{i+1})+l_i-l_{i-1} \\
  &=(\delta^{-d}-1)[(l_n-l_2)\delta^{-(i-1)d} +\sum_{j=1}^{i-2}(l_j-l_{j+2})\delta^{-(i-j-1)d}] \\
  &+ \delta^{-d}(l_{i-1}-l_{i+1})+l_i-l_{i-1} \\
  &>(\delta^{-d}-1)[\delta\cdot\delta^{-(i-1)d} +\sum_{j=1}^{i-2}(\delta-\delta^{-d+1})\delta^{-(i-j-1)d}] \\
  & -\delta^{-d}(\delta^{-d+1}-\delta)+\delta \\
  &=(\delta^{-d}-1)\delta[\delta^{-(i-1)d} +\sum_{j=1}^{i-2}(1-\delta^{-d})\delta^{-(i-j-1)d}] \\
  & -\delta^{-d}(\delta^{-d+1}-\delta)+\delta \\
  &=(\delta^{-d}-1)\delta\cdot\delta^{-d} -\delta^{-d}(\delta^{-d+1}-\delta)+\delta \\
  &=\delta>0.
\end{align*}
Therefore, $\widetilde{l}_i>\widetilde{l}_{i-1}$ holds and the ordering after operation on row $i$ is:
\[
l_{i+1}<l_{i+2}<\dots<l_n<\widetilde{l}_1<\widetilde{l}_2<\dots<\widetilde{l}_i.
\]

When $i=n$, we have $\max_{j\neq i}l_j=\widetilde{l}_{n-1}$ and $\min_{j\neq i}l_j=\widetilde{l}_1$, resulting in  $\widetilde{l}_n=\delta^{-d}(\widetilde{l}_{n-1}-\widetilde{l}_{1})+l_n$. Similarly,
\begin{align*}
  &\widetilde{l}_n-\widetilde{l}_{n-1}=(l_n-l_2)(\delta^{-nd}-\delta^{-(n-1)d})  \\
  &+\sum_{j=1}^{n-2}(l_j-l_{j+2})(\delta^{-(n-j)d}-\delta^{-(n-j-1)d}) \\
  &-\delta^{-2d}(l_{n}-l_{2})+\delta^{-d}(l_{n-1}-l_{1})+l_n-l_{n-1} \\
  &=(l_n-l_2)(\delta^{-nd}-\delta^{-(n-1)d}-\delta^{-2d}) \\
  &+\sum_{j=1}^{n-2}(l_j-l_{j+2})(\delta^{-(n-j)d}-\delta^{-(n-j-1)d}) \\
  &+\delta^{-d}(l_{n-1}-l_{1})+l_n-l_{n-1} \\
  &>(\delta)(\delta^{-nd}-\delta^{-(n-1)d}-\delta^{-2d}) \\
  &+\sum_{j=1}^{n-2}(\delta-\delta^{-d+1})(\delta^{-(n-j)d}-\delta^{-(n-j-1)d}) \\
  &+\delta^{-d}(\delta)+\delta \\
  &=\delta>0.
\end{align*}
The last inequation holds when $\delta^{-nd}-\delta^{-(n-1)d}-\delta^{-2d}>0$, which is correct for $n>2$ and small enough $\delta$.

After $n$ operations, we have $\widetilde{l}_1<\widetilde{l}_2<\dots<\widetilde{l}_n$. Note that:
\begin{align*}
    \widetilde{l}_n
    & = (l_n-l_2)\delta^{-nd}+\sum_{j=1}^{n-2}(l_j-l_{j+2})\delta^{-(n-j)d}\\
    &-(l_n-l_2)\delta^{-2d}+(l_{n-1}-l_1)\delta^{-d}+l_n\\
    &<(\delta^{-d+1}-\delta)\delta^{-nd}+(\delta^{-d+1}-\delta)\delta^{-d}+(\delta^{-d+1}-\delta) \\
    &=\delta(\delta^{-d}-1)(\delta^{-nd}+\delta^{-d}+1),
\end{align*}
thus $\widetilde{l}_i$ has the upper bound (denoted as $\Delta_h$).

To ensure that all tokens are
distinct, we will add two additional layers of the form $\boldsymbol{Z}\rightarrow \boldsymbol{Z}+(\Delta_h/\delta)\psi(\boldsymbol{Z};0)$.

\textbf{First Global Shift} Since $0<\widetilde{l}_1<\widetilde{l}_2<\dots<\widetilde{l}_n<\Delta_h$, the additional layer adds $(\Delta_h/\delta)(\max_{j\neq i}\boldsymbol{Z}_{j,:}\boldsymbol{u})\boldsymbol{e}^{(1)\top}$ for each $i$. Thus,
\[
\widetilde{l}^+_i=
\begin{cases}
  \widetilde{l}_i+(\Delta_h/\delta)\widetilde{l}_n & \text{if}\;i\neq n,\\
  \widetilde{l}_n+(\Delta_h/\delta)\widetilde{l}_{n-1} & \text{if}\;i=n.
\end{cases}
\]
For any $i, j\neq n$, we have $\widetilde{l}^+_i<\widetilde{l}^+_j$ if $i<j$. Note that:
\begin{align*}
    \widetilde{l}^+_1-\widetilde{l}^+_n
    &= (\Delta_h/\delta)(\widetilde{l}_n-\widetilde{l}_{n-1})+\widetilde{l}_1-\widetilde{l}_n \\
    &>(\Delta_h/\delta)\cdot\delta-\Delta_h\\
    &>0,
\end{align*}
the order after first global shift is
\[
    \widetilde{l}^+_n<\widetilde{l}^+_1<\widetilde{l}^+_2<\dots<\widetilde{l}^+_{n-1}.
\]

\textbf{Second Global Shift}
At the second global shift, we have
\[
\widetilde{l}^{++}_i=
\begin{cases}
  \widetilde{l}_i^++(\Delta_h/\delta)\widetilde{l}_{n-1}^+ & \text{if}\;i\neq n-1,\\
  \widetilde{l}_{n-1}^++(\Delta_h/\delta)\widetilde{l}_{n-2}^+ & \text{if}\;i=n-1.
\end{cases}
\]
By expansion, $\widetilde{l}^{++}_i$ has a more clear form as follows.
{\small
\begin{align*}
&\widetilde{l}^{++}_i \\
&=\begin{cases}
  \widetilde{l}_{n-1}+(\Delta_h/\delta)(\widetilde{l}_{n-2}+\widetilde{l}_{n})+(\Delta_h/\delta)^2\widetilde{l}_{n} & \text{if}\;i=n-1, \\
  \widetilde{l}_{n}+2(\Delta_h/\delta)\widetilde{l}_{n-1}+(\Delta_h/\delta)^2\widetilde{l}_n & \text{if}\;i=n, \\
  \widetilde{l}_i+(\Delta_h/\delta)(\widetilde{l}_{n-1}+\widetilde{l}_{n})+(\Delta_h/\delta)^2\widetilde{l}_{n} & \text{otherwise}.
\end{cases}
\end{align*}}

The output of the second global shift is our $g_c$ (i.e., $\widetilde{l}^{++}_i=g_c(\boldsymbol{G})_{i,:}\boldsymbol{u}$).
Finally, we verify two properties of contextual mapping in Definition~\ref{def:1}.

\begin{itemize}
    \item For any $\boldsymbol{G}\in G_\delta$, we have $g_c(\boldsymbol{G})_{i,:}\boldsymbol{u} \mod (\Delta_h/\delta)=\widetilde{l}_i$. All entries of $q(\boldsymbol{G})\boldsymbol{u}$ are distinct because $\widetilde{l}_i$ is distinct with each other.
    \item For any $\boldsymbol{G}_1, \boldsymbol{G}_2 \in G_\delta$ ($\boldsymbol{G}_1\neq \boldsymbol{G}_2$), each entry of $g_c(\boldsymbol{G}_i)\boldsymbol{u}$ lies in the interval $[(\Delta_h/\delta)^2\widetilde{l}_n, (\Delta_h/\delta)^2(\widetilde{l}_n+\delta))$. Since $\widetilde{l}_n$ is the unique identity for the input $\boldsymbol{G}$, all entries of $q(\boldsymbol{G}_1)$ and $q(\boldsymbol{G}_2)$ are distinct.
\end{itemize}

Therefore, $g_c(\boldsymbol{G})$ satisfies the definition of contextual mapping.
\end{proof}

\subsubsection{Value Mapping (Feed-forward)}
\begin{lemma}[Lemma 7 \citep{yun2019Transformers}]
  There exists a function $g_v$ composed of $n(1/\delta)^{dn}$ token-wise feed-forward layers with $r=1$ and piece-wise linear functions (at most three pieces), such that $g_v$ is defined by a token-wise function $g_v^{tkn}$,
  \[
  g_v(\boldsymbol{Z})= [g_v^{tkn}(\boldsymbol{Z}_1) \dots g_v^{tkn}(\boldsymbol{Z}_n)],
  \]
  where
  \[
  g_v^{tkn}(\boldsymbol{Z}_i)=g_v^{tkn}(g_c(\boldsymbol{G})_i)=f(\boldsymbol{G}_i).
  \]
\end{lemma}

Therefore, we have $\bar{g}(\boldsymbol{X})=g_v\circ g_c\circ g_q(\boldsymbol{X})=\bar{f}(\boldsymbol{X})$ expect for a set has measure $O(\delta^{d})$ \citep{yun2019Transformers}, which implies that $d_p(\bar{f},\bar{g})\leq O(\delta^{d/p})$.

\subsection{Proof for Step 3}
\begin{lemma}[Lemma 9 \citep{yun2019Transformers}]
For each modified Transformer blocks $\bar{g}\in \bar{\mathcal{T}}^{2,1,1}$, there exists the
Transformer without diag-attention blocks $g\in \mathcal{T}^{2,1,4}$ such that  $d_p(\bar{g},g) \leq \epsilon/3$.
\end{lemma}

Since the modification is only the softmax function and ReLU activation function (not related with the self-attention matrix $A$), the lemma still holds.

By Summarizing the above three steps, we have:
\[
d_p(f,g)\leq d_p(f,\bar{f})+d_p(\bar{f},\bar{g})+d_p(\bar{g},g)\leq 2\epsilon/3+O(\delta^{d/p}).
\]
With enough small $\delta$, we have $d_p(f,g) \leq \epsilon$.
Thus, Transformers
without diag-attention are also universal approximators.

\clearpage
\section{Data Set \label{app:dataset}}
\subsection{MNLI}
The Multi-Genre Natural Language Inference \cite{williams2018broad} is a crowdsourced ternary classification task. Given a premise sentence and a hypothesis sentence, the target is to predict whether the last sentence is an [entailment], [contradiction], or [neutral] relationships with respect to the first one.

\subsection{QQP}
The Quora Question Pairs \cite{chen2018quora} is a binary classification task. Given two questions on Quora, the target is to determine whether these two asked questions are semantically equivalent or not.

\subsection{QNLI}
The Question Natural Language Inference \cite{wang2018multi} is a binary classification task derived from the Stanford Question Answering Dataset \cite{rajpurkar2016squad}. Given sentence pairs (question, sentence), the target is to predict whether the last sentence contains the correct answer to the question.

\subsection{SST-2}
The Stanford Sentiment Treebank \cite{socher2013recursive} is a binary sentiment classification task for a single sentence. All sentences are extracted from movie reviews with human annotations of their sentiment.

\subsection{CoLA}
The Corpus of Linguistic Acceptability \cite{warstadt2019neural} is a binary classification task consisting of English acceptability judgments extracted from books and journal articles. Given a single sentence, the target is to determine whether the sentence is linguistically acceptable or not.

\subsection{STS-B}
The Semantic Textual Similarity Benchmark \cite{cer2017semeval} is a regression task for predicting the similarity score (from $1$ to $5$) between a given sentence pair, whose sentence pairs are drawn from news headlines and other sources.

\subsection{MRPC}
The Microsoft Research Paraphrase Corpus \cite{dolan2005automatically} is a binary classification task. Given a sentence pair extracted from online news sources, the target is to determine whether the sentences in the pair are semantically equivalent.

\subsection{RTE}
The Recognizing Textual Entailment \cite{bentivogli2009fifth} is a binary entailment classification task similar to MNLI, where [neutral] and [contradiction] relationships are classified into [not entailment].

\subsection{SWAG}
The Situations with Adversarial Generations \cite{zellers2018swag} is a multiple-choice task consisting of $113$K questions about grounded situations. Given a source sentence, the task is to select the most possible one among four choices for sentence continuity.

\subsection{SQuAD v1.1}
The Stanford Question Answering Dataset (SQuAD v1.1) \cite{rajpurkar2016squad} is a large-scale question and answer task consisting of $100$K question and answer pairs from more than $500$ articles. Given a passage and the question from Wikipedia, the goal is to determine the start and the end token of the answer text.

\subsection{SQuAD v2.0}
The SQuAD v2.0 task \cite{rajpurkar2018know} is the extension of above SQuAD v1.1, which contains the $100$K questions in SQuAD v1.1 and $50$K unanswerable questions. The existence of unanswerable question makes this task more realistic and challenging.

\section{Implementation Details \label{app:hyper}}
The hyper-parameters of various downstream tasks are shown in Table~\ref{tbl:hyper}.

\begin{table*}[b]
\begin{center}
\caption{Hyper-parameters for different downstream tasks. \label{tbl:hyper}}
\resizebox{0.65\textwidth}{!}{
\begin{tabular}{lcccc}
\hline
          & GLUE & SWAG &SQuAD v1.1 & SQuAD v2.0  \tabularnewline
\hline
Batch size & 32 & 16 & 32 & 48  \tabularnewline
Weight decay & [0.1, 0.01]& [0.1, 0.01] &[0.1, 0.01] & [0.1, 0.01] \tabularnewline
Warmup proportion & 0.1 & 0.1 & 0.1& 0.1 \tabularnewline
Learning rate decay & linear & linear &linear & linear \tabularnewline
Training Epochs & 3 & 3 & 3 &2 \tabularnewline
Learning rate & \multicolumn{4}{c}{[2e-5, 1e-5, 1.5e-5, 3e-5, 4e-5, 5e-5]} \tabularnewline
\hline
\end{tabular}}
\end{center}
\label{tab:implementation_details}
\end{table*}

%%%%%%%%%%%%%%%%%%%%%%%%%%%%%%%%%%%%%%%%%%%%%%%%%%%%%%%%%%%%%%%%%%%%%%%%%%%%%%%
%%%%%%%%%%%%%%%%%%%%%%%%%%%%%%%%%%%%%%%%%%%%%%%%%%%%%%%%%%%%%%%%%%%%%%%%%%%%%%%
\end{document}